\documentclass[pdflatex,sn-mathphys-num]{sn-jnl}

\usepackage{graphicx}%
\usepackage{multirow}%
\usepackage{amsmath,amssymb,amsfonts}%
\usepackage{amsthm}%
\usepackage{mathrsfs}%
\usepackage[title]{appendix}%
\usepackage{xcolor}%
\usepackage{textcomp}%
\usepackage{manyfoot}%
\usepackage{booktabs}%
\usepackage{array}
\usepackage{algorithm}%
\usepackage{algorithmicx}%
\usepackage{algpseudocode}%
\usepackage{listings}%

\usepackage{float}
\usepackage{rotating}
\usepackage{makecell}
\usepackage{subcaption} 
\usepackage{graphicx}
\usepackage{booktabs}
\usepackage{amsmath}
\usepackage{cleveref} 
\usepackage{amsmath}
\usepackage{makecell}
\usepackage{subcaption} 

\usepackage[
    width=0.9\textwidth,
    font=small,
    labelfont=bf,
    labelsep=period,
    justification=justified,
    singlelinecheck=true 
]{caption}

\usepackage{tikz}
\usetikzlibrary{shapes.geometric, arrows.meta, shadows, positioning, calc}
\usepackage{adjustbox} 

\theoremstyle{thmstyleone}%
\theoremstyle{thmstyletwo}%

\theoremstyle{thmstylethree}%

\begin{document}

\title[Article Title]{Importance inversion transfer identifies shared principles for cross-domain learning}


\author*[1,2,3]{\fnm{Daniele} \sur{Caligiore}}\email{daniele.caligiore@istc.cnr.it}

\affil*[1]{\orgdiv{Computational and Translational Neuroscience Laboratory, Institute of Cognitive Sciences and Technologies}, \orgname{National Research Council (CTNLab-ISTC-CNR)}, \orgaddress{\street{Via Gian Domenico Romagnosi, 18A}, \city{Rome}, \postcode{00196}, \country{Italy}}}

\affil[2]{\orgname{AI2Life s.r.l., Innovative Start-Up, ISTC-CNR Spin-Off,}, \orgaddress{\street{Via Sebino, 32}, \city{Rome}, \postcode{00199}, \country{Italy}}}

\affil[3]{\orgdiv{Department of Human Sciences, Communication, Education and Psychology}, \orgname{Libera Università Maria Ss. Assunta (LUMSA)}, \orgaddress{\street{Via della Traspontina, 21}, \city{Rome}, \postcode{00193}, \country{Italy}}}


\abstract{The capacity to transfer knowledge across scientific domains relies on shared organizational principles. However, existing transfer-learning methodologies often fail to bridge radically heterogeneous systems, particularly under severe data scarcity or stochastic noise. This study formalizes \emph{Explainable Cross-Domain Transfer Learning} (X-CDTL), a framework unifying network science and explainable artificial intelligence to identify structural invariants that generalize across biological, linguistic, molecular, and social networks. By introducing the \emph{Importance Inversion Transfer} (IIT) mechanism, the framework prioritizes domain-invariant structural anchors over idiosyncratic, highly discriminative features. In anomaly detection tasks, models guided by these principles achieve significant performance gains---exhibiting a \emph{56\% relative improvement in decision stability under extreme noise}---over traditional baselines. These results provide evidence for a shared organizational signature across heterogeneous domains, establishing a principled paradigm for cross-disciplinary knowledge propagation. By shifting from opaque latent representations to explicit structural laws, this work advances machine learning as a robust engine for scientific discovery.}

\keywords{Complex systems, Explainable Cross-Domain Transfer learning (X-CDTL), Domain adaptation, Network science, Explainable Machine Learning, Topological invariants, Importance Inversion Transfer (IIT), Anomaly detection}



\maketitle

\section{Introduction}
\label{sec:intro}

Scientific progress increasingly necessitates the synthesis of knowledge across domains that differ radically in scale, modality, and underlying mechanisms. Meaningful cross-disciplinary propagation---ranging from leveraging biological analogies to predict engineering failures to inferring linguistic organization from social networks---presupposes shared organizational principles \cite{nesse2025explanations, li2025application}. Identifying these principles is not merely a descriptive exercise but a fundamental prerequisite for principled generalization and robust knowledge transfer.

Transfer learning (TL) facilitates the adaptation of predictive functions across disparate datasets by leveraging commonalities in feature representations or underlying generative processes \cite{pan2009survey, weiss2016survey}. This paradigm shift from isolated learning to knowledge propagation addresses the inherent limitations of standard machine learning when facing novel domains with distinct probability distributions. However, domain adaptation theory posits that the error bound on a target task remains strictly contingent upon the divergence between source and target distributions \cite{mansour2008domain, ben2010theory, ganin2016domain}. In real-world scientific applications, stochastic corruption and feature noise exacerbate this divergence, frequently causing standard alignment methods to collapse and necessitating more robust approaches to isolate domain-invariant representations.

Furthermore, existing transfer methodologies primarily target closely related settings where datasets share similar generative dynamics \cite{pan2009survey, gholizade2025review}. When applied across fundamentally heterogeneous systems, conventional latent alignment techniques often yield uninterpretable, domain-specific embeddings that obscure the mechanistic pathways of knowledge propagation. This lack of transparency hinders the identification of the structural properties that consistently connect disparate domains \cite{raghu2019transfusion, zhao2019learning}.

Network science represents a powerful abstraction for cross-disciplinary synthesis, as entities and interactions in systems ranging from molecular graphs to social structures can be mapped onto complex networks \cite{milo2002network, posfai2016network}. By representing interactions as nodes and edges, this approach demonstrates extensive utility across fields as diverse as materials science, cosmology, and systems biology, enabling the analysis of fundamental phenomena such as phase transitions, information diffusion, and technological innovation \cite{iniguez2020bridging, das2023key, Patel_Sinha_Palermo_2024, Choudhary_2021, Batzner_2022, Aroboto-APL-2023, batatia2025, strey2023graph, artime2024robustness, rosato2008is}. 

However, the current reliance on handcrafted descriptors often fails to distinguish functionally meaningful invariants from artifacts of sampling or idiosyncratic domain constraints, particularly when faced with noisy or scarce datasets \cite{artzy2004comment, broido2019scale}. To resolve this impasse, the present study formalizes \emph{Explainable Cross-Domain Transfer Learning} (X-CDTL), a paradigm unifying network science with explainable artificial intelligence (XAI). Building upon the theoretical foundations established in \cite{caligiore2025exploring}, this framework facilitates the identification of shared structural principles that remain invariant across disparate disciplines. Through this lens, complex systems with radically different generative mechanisms become comparable within a standardized topological space, supporting robust functional transfer under realistic experimental constraints.

The framework prioritizes Explainable Machine Learning (XML) architectures to ensure transparency and systematic trust \cite{angelini2024unraveling, damore2024explainable, islam2022systematic, linardatos2020explainable}. Beyond providing interpretable decision pathways, the capacity of XML to quantify feature importance serves as a critical diagnostic tool for cross-domain comparison. In this context, explainability functions not merely as a validation layer but as an active discovery mechanism \cite{caligiore2025exploring, rudin2019stop}: feature importance values act as quantitative proxies for the functional relevance of specific organizational principles.

Rather than aligning opaque latent representations, X-CDTL identifies and transfers explicit structural principles governing system organization. These principles underpin a novel strategy termed \emph{Importance Inversion Transfer} (IIT). In a significant departure from conventional feature selection \cite{guyon2003introduction, meinshausen2010stability, hooker2021unrestricted}, IIT leverages XML to isolate topological descriptors with the \textit{lowest} discriminative utility for domain classification. By prioritizing these stable, domain-invariant structural anchors, the framework filters out idiosyncratic noise in favor of fundamental topological properties. 

Validation across diverse real-world networks reveals a compact set of features, including assortativity and path efficiency, that consistently capture structural invariants. 
These structural anchors demonstrate significant functional relevance in demanding transfer learning scenarios, particularly for anomaly detection under extreme signal degradation. In high-noise regimes, models guided by these common descriptors achieve a substantial \textit{rescue effect}, exhibiting up to a 56\% relative improvement in decision stability compared to no-transfer baselines. The persistence of these gains under maximum feature corruption confirms that the identified topological pillars encode transferable organizational laws rather than domain-specific artifacts, effectively regularizing the decision manifold when local structural signals are most severely compromised.

Empirical validation demonstrates that cross-domain knowledge propagation achieves maximal efficacy when anchored to statistically stable structural invariants. This finding uncovers a \emph{transfer paradox}, where generalization performance peaks at intermediate levels of structural similarity rather than total isomorphism. By synthesizing predictive precision with explanatory depth, the X-CDTL framework transforms machine learning from a descriptive utility into a discovery engine capable of elucidating the shared principles governing complex systems. This paradigm provides a principled foundation for interpretable generalization, establishing a robust framework for rigorous scientific inquiry across disparate disciplinary boundaries.

\section{Results}
\label{sec:results}

\subsection{Structural Characterization and Topological Fingerprints}

\begin{figure}[h!]
    \centering
    \begin{subfigure}{0.48\textwidth}
        \includegraphics[width=\textwidth]{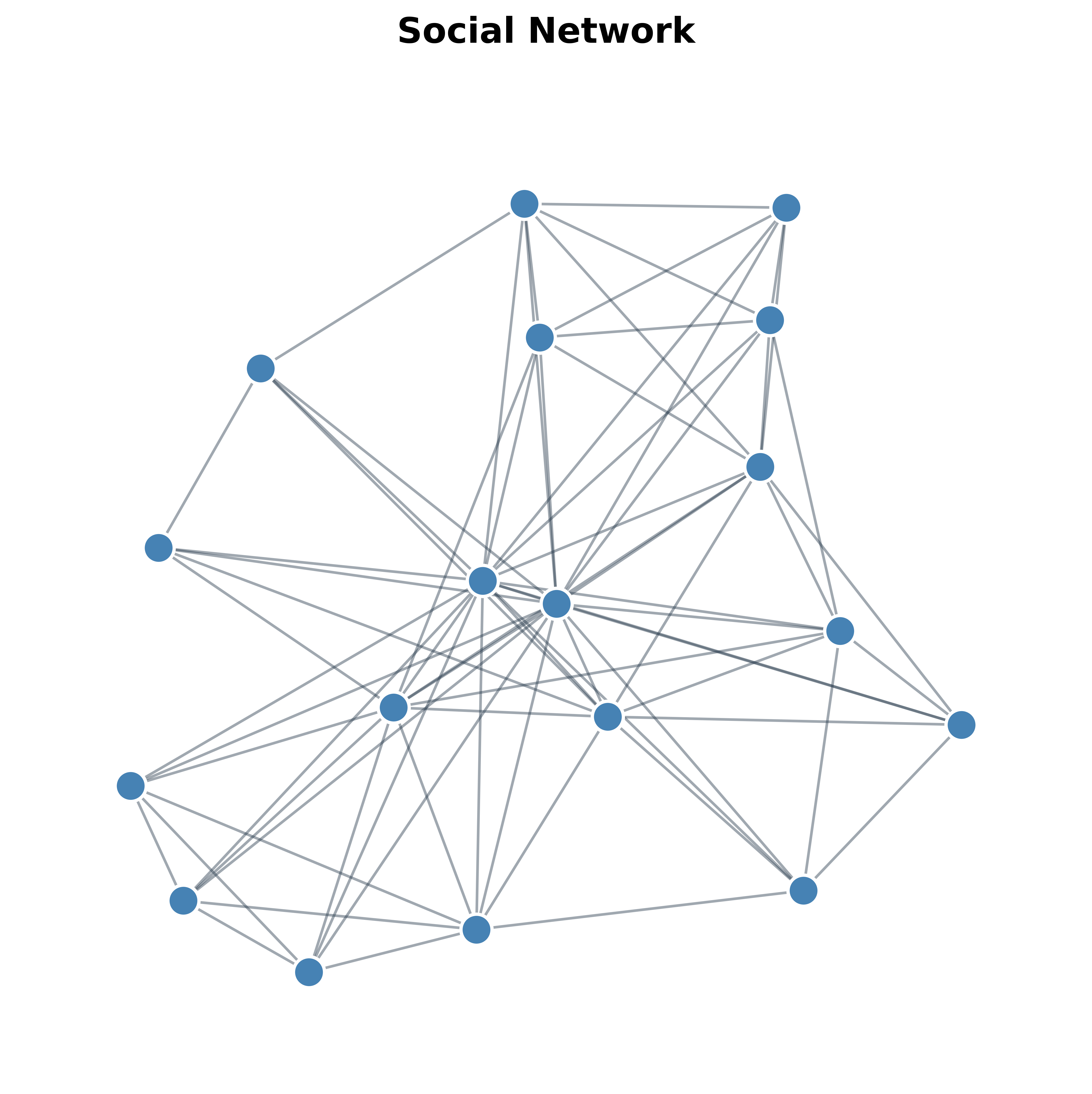}
    \end{subfigure}
    \hfill
    \begin{subfigure}{0.48\textwidth}
        \includegraphics[width=\textwidth]{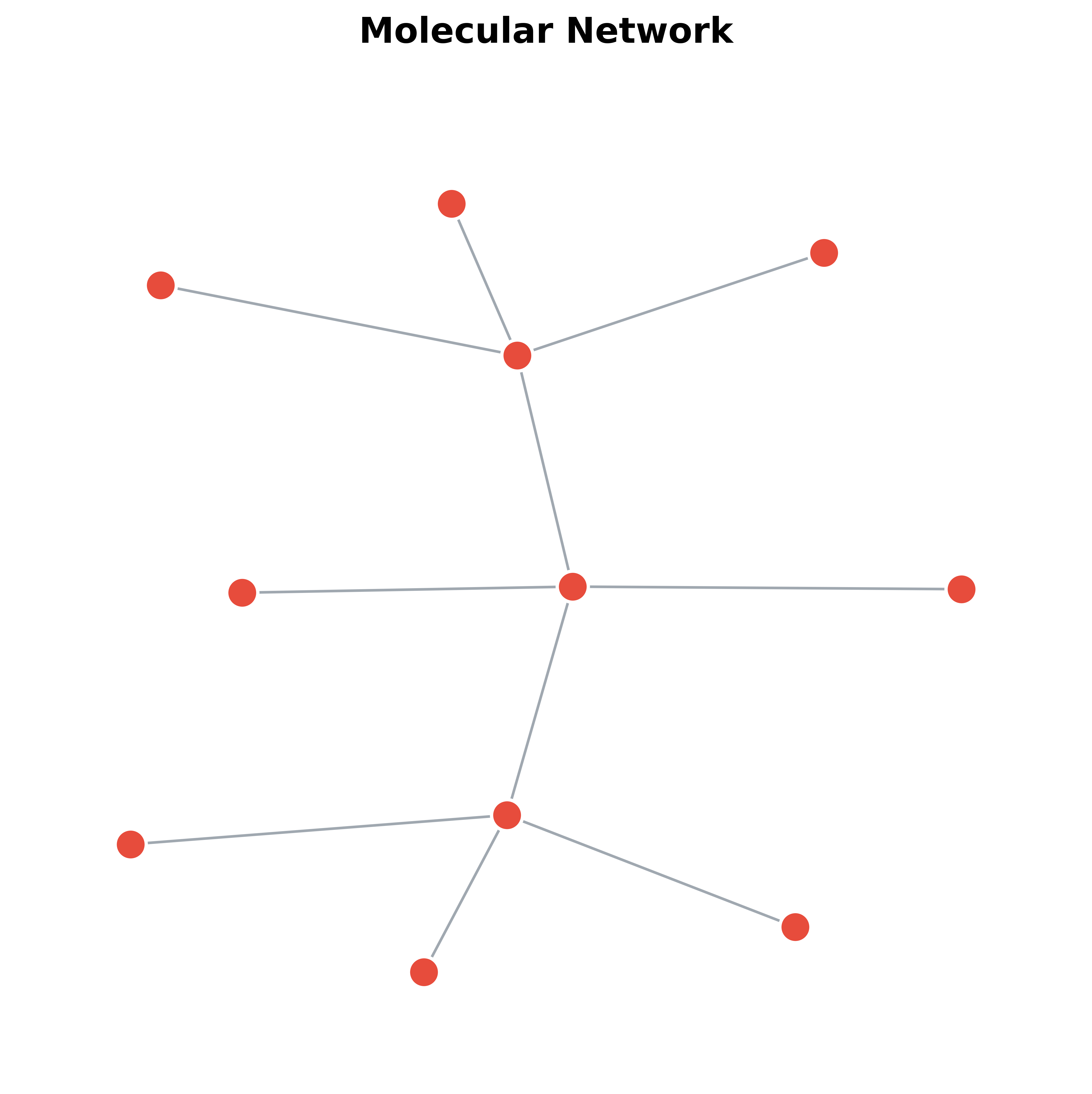}
    \end{subfigure}
    \vspace{10pt} 
    \begin{subfigure}{0.48\textwidth}
        \includegraphics[width=\textwidth]{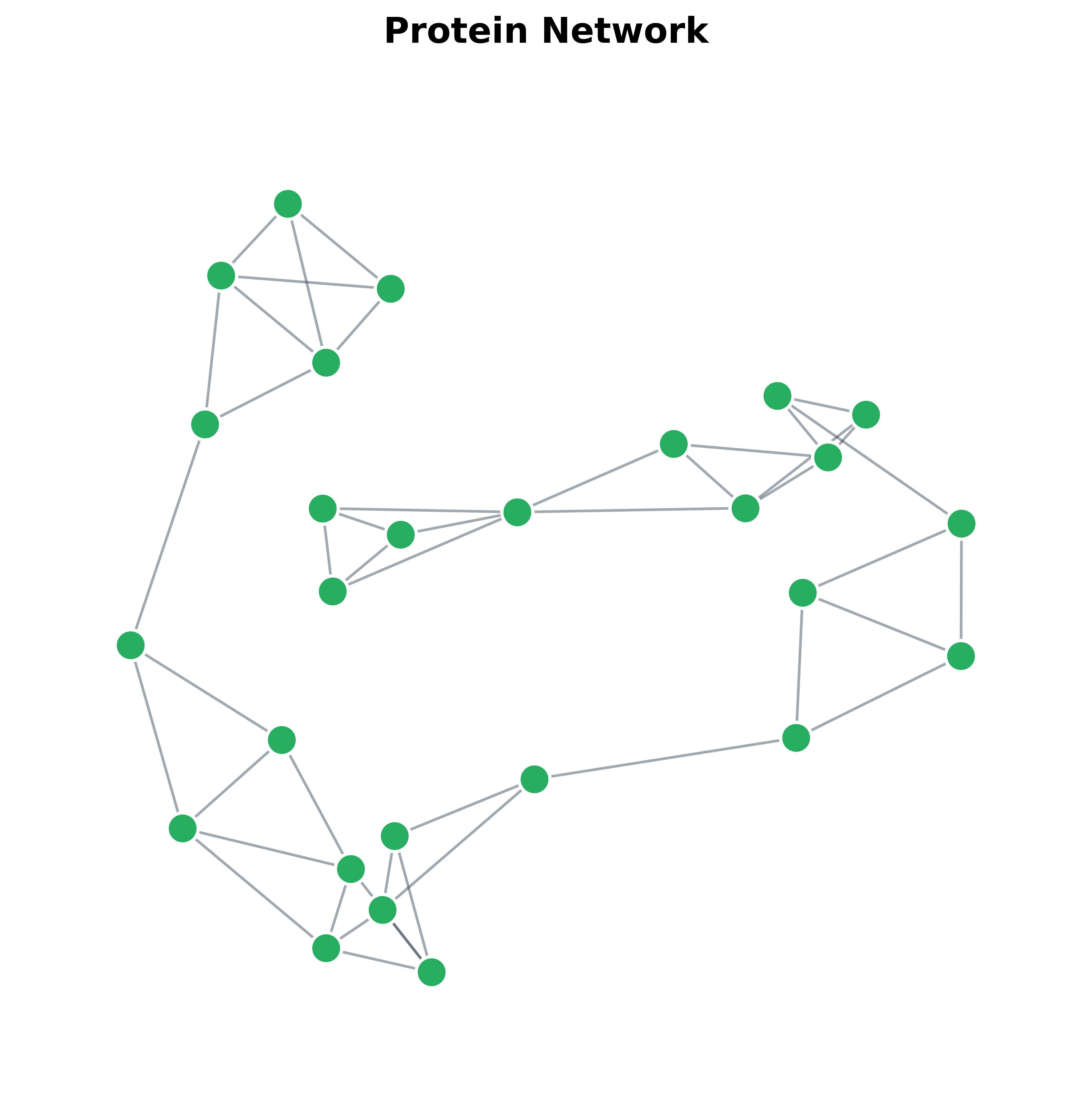}
    \end{subfigure}
    \hfill
    \begin{subfigure}{0.48\textwidth}
        \includegraphics[width=\textwidth]{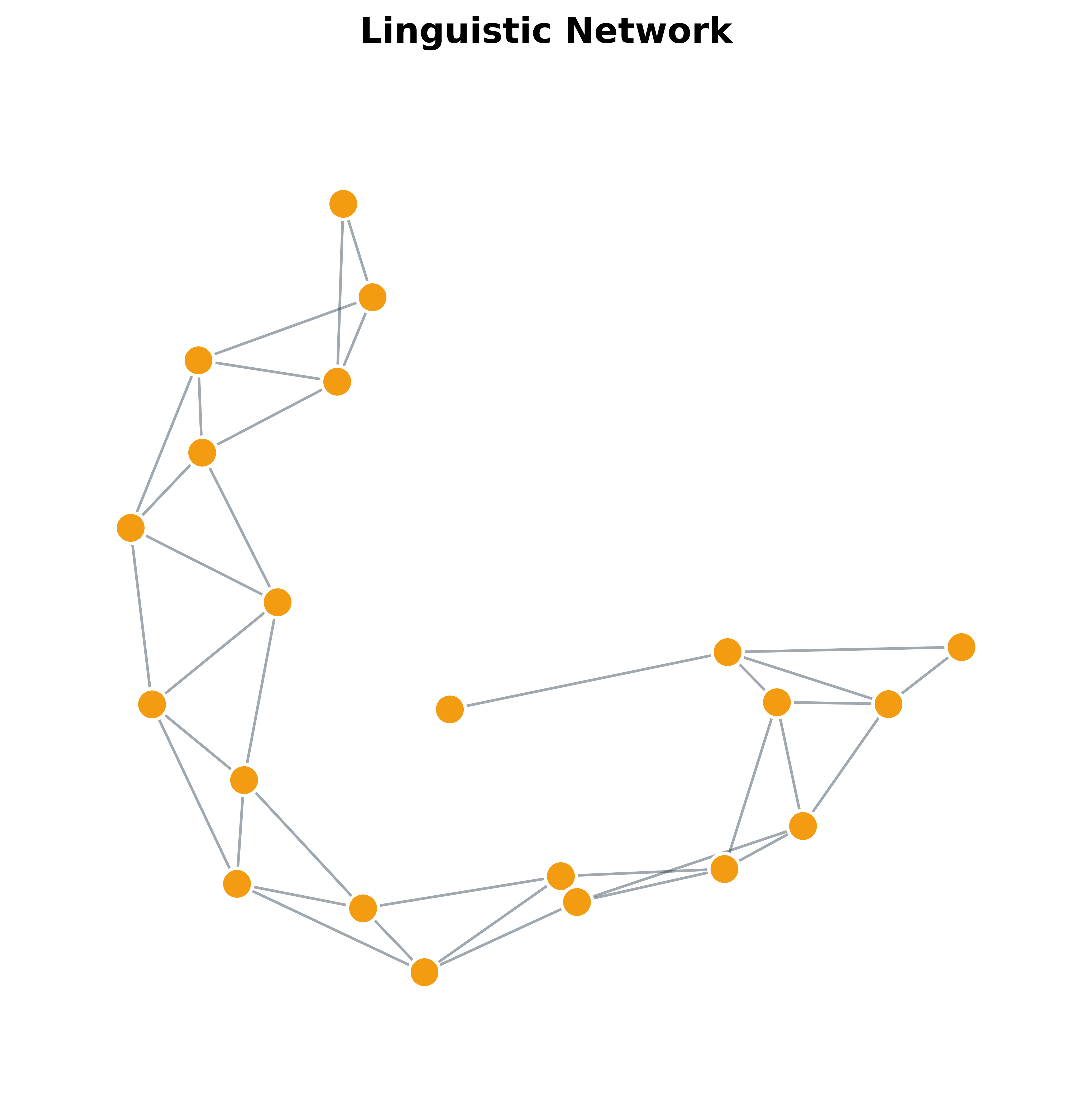}
    \end{subfigure}
    \caption{\textbf{Structural landscapes and topological diversity across domains.} Representative graph samples illustrate distinct density, modularity, and branching patterns. Connectivity definitions vary across scientific scales: in social networks, nodes represent users connected by friendships; in molecular graphs, nodes denote atoms joined by chemical bonds; in protein networks, nodes represent amino acids linked by physical interactions; and in linguistic networks, nodes denote words connected by contextual co-occurrence. These architectures underpin the X-CDTL framework by providing a heterogeneous set of structural priors for the manifold alignment pipeline.}
    \label{fig:example_graphs}
\end{figure}

The structural foundation of the X-CDTL framework relies on a multi-scale characterization of four distinct graph domains---social ego-networks, molecular graphs, protein interaction networks, and linguistic co-occurrence networks. Representative topologies (Fig.~\ref{fig:example_graphs}) reveal the diverse connectivity regimes involved, ranging from the ultra-dense, small-world architectures of social ego-networks to the sparse, valency-constrained layouts of molecular graphs. These structural fingerprints provide the morphological basis for evaluating the framework generalization capacity across fundamentally different generative dynamics.

A set of twelve topological descriptors, spanning connectivity, clustering, spectral, and modular dimensions, quantifies this diversity (ensemble statistics in Supplementary Table~\ref{tab:supp_descriptive_stats}). The selected domains occupy non-overlapping topological regimes, as demonstrated by the high separability in the structural feature space (Supplementary Fig.~\ref{fig:boxplots_supplementary}). 
Social networks exhibit quasi-clique architectures characterized by high local redundancy, with a mean clustering coefficient of $0.84 \pm 0.06$ and a density ($0.65 \pm 0.17$) significantly exceeding all other domains. Conversely, molecular graphs constitute a sparse, nearly acyclic space where connectivity is governed by chemical valence constraints, resulting in near-zero average clustering ($0.01 \pm 0.03$). Protein and linguistic networks occupy intermediate regimes; notably, protein networks are uniquely distinguished by high modularity ($0.52 \pm 0.11$), reflecting the hierarchical community organization essential for biological function.

Spectral fingerprints further emphasize these disparities. The spectral radius of social networks ($12.28 \pm 4.32$) is approximately five times larger than that of molecular systems ($2.56 \pm 0.21$), indicating vast differences in global connectivity and information propagation potential. Despite the intrinsic variability inherent to real-world relational data, the interquartile ranges of key features—specifically Density and Spectral Radius—remain non-overlapping across the majority of domain pairs. This robust separation confirms that the selected domains are topologically distinct yet internally consistent, providing a demanding benchmark for cross-domain alignment.

\subsection{Structural Encoding and Domain Separability}

\subsubsection{Topological Discriminative Power and Benchmarking}

Topological divergence across domains establishes structural metrics as high-fidelity fingerprints of network identity. A comprehensive benchmarking analysis quantifies this encoding capacity by evaluating the separability of the four network classes---\textit{social, molecular, protein,} and \textit{linguistic}---using an XML framework built upon standardized topological descriptors.

Performance evaluation across three distinct learning architectures (\textit{Gradient Boosting, Logistic Regression,} and \textit{Random Forest}) confirms a consistently high discriminative utility. As reported in Table~\ref{tab:performance}, metrics marginalized over 10 independent random seeds demonstrate classification accuracies within the narrow range of $[96.5\%,\,96.9\%]$, supported by near-perfect ROC-AUC values ($\geq 0.996$). These results indicate that the structural feature space is intrinsically well-organized, yielding nearly optimal decision boundaries that remain robust even under stochastic resampling of the network ensembles.

\begin{table}[h]
\centering
\caption{\textbf{Domain classification performance across learning architectures.} Metrics represent ensemble averages over 10 independent random seeds ($N_{\text{total}} = 20,000$ networks). The consistently high discriminative power indicates that standardized topological features provide a robust, domain-invariant signal sufficient for high-fidelity network identification.}
\label{tab:performance}
\small
\renewcommand{\arraystretch}{1.2}
\begin{tabular}{lccc}
\toprule
\textbf{Model} & \textbf{Accuracy} & \textbf{F1-macro} & \textbf{ROC-AUC} \\
\midrule
Gradient Boosting      & 0.969 & 0.968 & 0.997 \\
Logistic Regression    & 0.965 & 0.965 & 0.996 \\
Random Forest          & 0.965 & 0.965 & 0.997 \\
\bottomrule
\end{tabular}
\end{table}

The marginal performance gap (approximately 0.3\%) between the top-performing Gradient Boosting model and the linear Logistic Regression baseline is theoretically significant. Notably, the fact that a linear model achieves performance parity with complex ensemble methods such as Random Forest suggests that the morphological signatures defining network domains are not only expressive but also linearly accessible within the standardized feature space. Such results validate the existence of domain-specific \emph{structural regimes}, in which specific collections of topological descriptors converge to define highly separable morphological manifolds. Further classification diagnostics, including the ensemble confusion matrix for the Gradient Boosting model, are detailed in the Supplementary Information (Supplementary Fig.~\ref{fig:fig_conf_mat_supp}). 

\subsubsection{Topological Backbone and Global Consensus Ranking}

The identification of structural anchors is governed by the hierarchical aggregation of discriminative utility, rank-order consistency, and metric stability. 
While supervised classification diagnostics identify the primary discriminants for domain separation (see Supplementary Table~\ref{tab:borda_importance}), the framework formalizes the transition from domain-specific importance to general transferability via the \emph{Global Consensus IIT$_{\text{score, G}}$}. For each topological descriptor, this index represents the mean of its directed $\text{IIT}_{\text{score}}$ values across all 12 source--target combinations, isolating the structural similarities capable of bridging disparate generative dynamics.

\begin{figure}[H]
    \centering
    \includegraphics[width=0.7\textwidth]{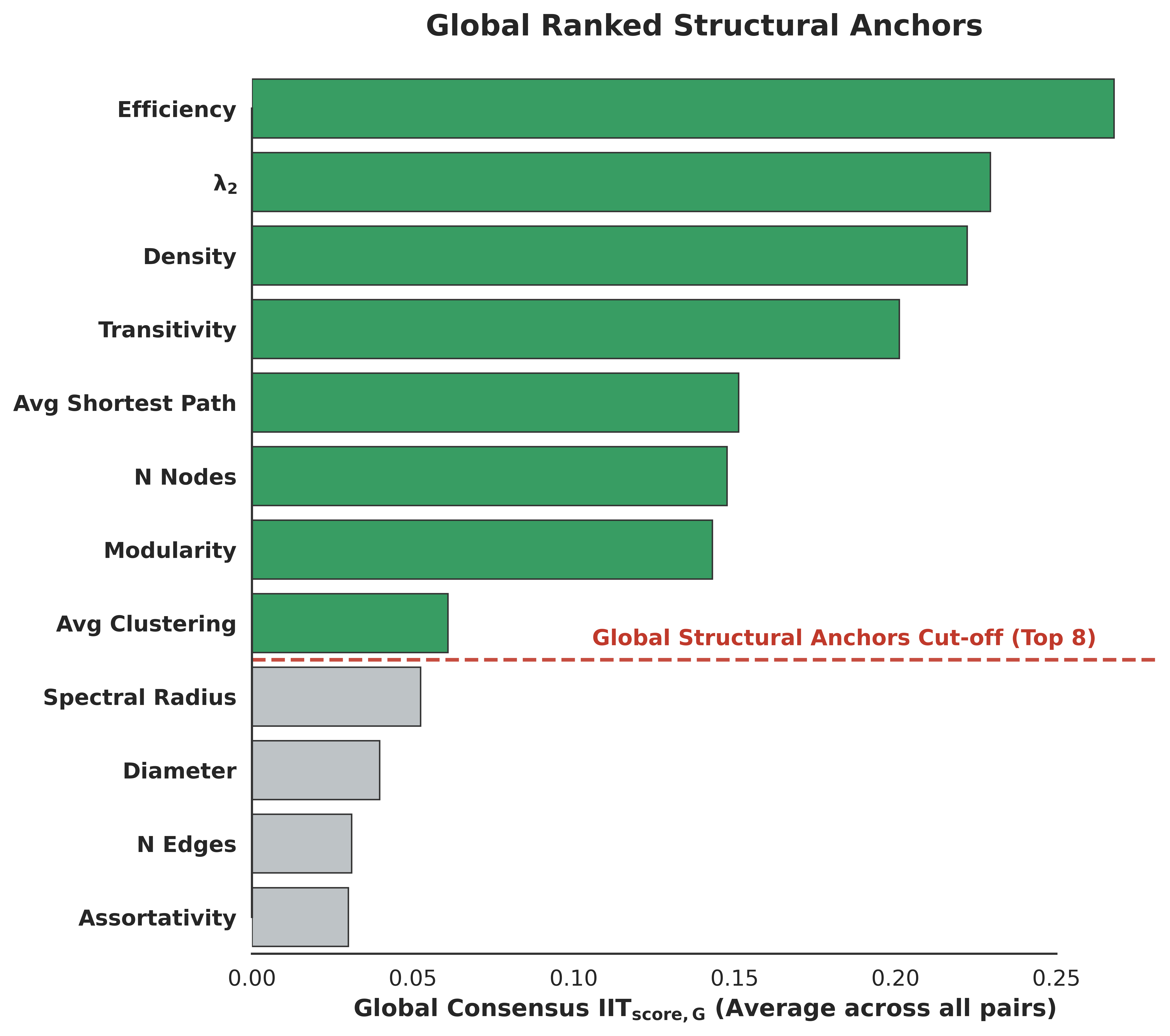}
    \caption{\textbf{Global ranking of structural anchors.} Hierarchy of the topological feature space based on the Global Consensus $\text{IIT}_{\text{score, G}}$. The green bars identify the eight structural anchors. The hierarchy reveals a significant reordering of features compared to raw discriminative importance and identifies a distinct performance gap after the eighth descriptor, justifying the parsimonious 12/8 configuration. The red dashed line designates the cut-off boundary that isolates domain-invariant anchors from idiosyncratic, noise-prone markers.}
    \label{fig:global_ranking}
\end{figure}

The resulting hierarchy reveals a significant inversion of the feature importance profile (Fig.~\ref{fig:global_ranking}).
In the raw Borda ranking, descriptors such as the \textit{average clustering coefficient} (Borda score: 2.267), \textit{spectral radius} (4.233), and \textit{diameter} (4.333) emerge as the strongest discriminants for domain identification. However, the $\text{IIT}_{\text{score, G}}$ identifying these features as either highly volatile across scientific scales or as encoding information that is too domain-specific to facilitate alignment. Consequently, the consensus ranking demotes spectral and global-scale markers to the bottom of the transferability hierarchy.

Conversely, descriptors such as \textit{efficiency}, $\lambda_2$, and \textit{density}---which occupy intermediate ranks in terms of raw discriminative utility---ascend to the top of the global consensus, exhibiting the highest alignment potential across disparate manifolds. 
This reordering identifies a parsimonious set of eight \textit{structural anchors} (green bars in Fig.~\ref{fig:global_ranking}) that constitute the shared topological backbone of the framework. Within this regime, \textit{efficiency} and $\lambda_2$ exhibit superior stability, providing a robust \emph{structural grammar} for knowledge propagation.

While the $\text{IIT}_{\text{score, G}}$ establishes the global foundation for this strategy, the framework allows for a directed refinement of this backbone for specific source--target combinations. By accounting for pair-specific rank-order consistency and distributional proximity, this directed scoring enables the identification of \textit{local structural anchors}---descriptors that may exhibit global volatility at the ensemble level but possess significant metric compatibility for particular transitions. A salient instance is observed with \textit{modularity}, which re-emerges as a primary structural anchor for specific leaps (e.g., in transfers involving \textit{Molecular} or \textit{Social} domains) despite its intermediate global rank. 

By integrating these hierarchical indices, the selection process bypasses heuristic thresholds to prioritize features that maximize the trade-off between functional relevance and metric stability. Together, the $\text{IIT}_{\text{score, G}}$ and the directed $\text{IIT}_{\text{score}}$ ensure that the transfer process remains anchored to a domain-invariant structural backbone, effectively insulating knowledge propagation from the idiosyncratic noise identified in the descriptive statistical profiling. 

\subsubsection{The Landscape of Pairwise Feature Transferability}
\label{subsubsec:pairwise_transferability}

A directed pairwise investigation across all source--target combinations (Table~\ref{tab:pairwise_transfer}) provides a granular characterization of structural information propagation. In accordance with the hierarchical selection protocol (Sec.~\ref{subsubsec:steps}), the identification and ranking of these structural anchors are governed by the composite $\text{IIT}_{\text{score}}$. This criterion integrates (i) discriminative neutrality, (ii) rank-order consistency, and (iii) metric compatibility to determine the alignment potential between disparate manifolds. The reported Mean $\text{IIT}_{\text{score}}$ ($\overline{\text{IIT}}_{\text{score}}$) in Table~\ref{tab:pairwise_transfer} quantifies the aggregate strength of the structural bridge between domain pairs, ensuring that knowledge transfer is grounded in features that are both theoretically invariant and metrically stable.

\setlength{\rotFPtop}{0pt plus 1fil}
\setlength{\rotFPbot}{0pt plus 1fil}
\begin{sidewaystable*}[p]
\centering
\captionsetup{width=0.85\textwidth}
\caption{\textbf{Pairwise Transferable Features Summary.} Top 8 structural anchors for each domain pair, calculated over 10 independent seeds. Due to the observed symmetry in alignment potential, pairs are reported as bidirectional. Anchors are identified and ranked according to the composite $\text{IIT}_{\text{score}}$, ensuring a principled balance between low discriminative bias, rank-order consistency, and metric compatibility. The Mean $\text{IIT}_{\text{score}}$ ($\overline{\text{IIT}}_{\text{score}}$) represents the aggregate alignment potential for the selected feature subset. Pairs follow the taxonomic sequence: Social (Soc), Molecular (Mol), Proteins (Prot), and Linguistic (Ling).}
\label{tab:pairwise_transfer}
\footnotesize
\begin{tabular}{lcp{12cm}}
\toprule
\textbf{Domain Pair} & \textbf{$\overline{\text{IIT}}_{\text{score}}$} & \textbf{Structural Anchors} \\
\midrule
Soc $\leftrightarrow$ Mol   & 0.2111 & transitivity, $\lambda_2$, efficiency, avg\_shortest\_path, density, assortativity, modularity, diameter \\
Soc $\leftrightarrow$ Prot  & 0.1576 & efficiency, density, transitivity, $\lambda_2$, avg\_clustering, modularity, n\_nodes, assortativity \\
Soc $\leftrightarrow$ Ling  & 0.1430 & n\_nodes, efficiency, $\lambda_2$, transitivity, avg\_clustering, modularity, density, assortativity \\
\addlinespace
Mol $\leftrightarrow$ Prot  & 0.1638 & efficiency, density, $\lambda_2$, modularity, transitivity, n\_nodes, assortativity, avg\_shortest\_path \\
Mol $\leftrightarrow$ Ling  & 0.1794 & $\lambda_2$, efficiency, density, modularity, avg\_shortest\_path, assortativity, n\_nodes, transitivity \\
\addlinespace
Prot $\leftrightarrow$ Ling & 0.2480 & density, efficiency, n\_nodes, $\lambda_2$, transitivity, modularity, assortativity, n\_edges \\
\bottomrule
\end{tabular}
\end{sidewaystable*} 

The analysis of directed pairs reveals how domain-specific generative dynamics shape the nature of transferable structure. Notably, the \textit{Proteins $\leftrightarrow$ Linguistic} transfer exhibits the most significant structural homology ($\overline{\text{IIT}}_{\text{score}} = 0.2480$), driven primarily by \textit{density}, \textit{efficiency}, and \textit{n\_nodes}. This finding identifies a shared hierarchical organization common to information-bearing networks, where modular sub-structures (functional protein domains and syntactic word clusters) facilitate distributed information flow. 

A high degree of compatibility is also observed in the \textit{Social $\leftrightarrow$ Molecular} transfer ($\overline{\text{IIT}}_{\text{score}} = 0.2111$), characterized by \textit{transitivity}, \textit{algebraic connectivity} ($\lambda_2$), and \textit{efficiency}. This suggests that, despite profound differences in generative mechanisms, both systems share fundamental constraints on triangle closure and global connectivity resilience. In social systems, these properties arise from triadic closure and small-world effects, whereas in molecular graphs, they reflect chemical bonding constraints and spatial embedding. Their joint selection as top anchors indicates that transfer succeeds when domains share comparable balances between local cohesion and global integration.

Conversely, transfers between \textit{Social} and \textit{Linguistic} networks yield the most divergent results ($\overline{\text{IIT}}_{\text{score}} = 0.1430$). While social networks are driven by local reinforcement and community overlap, linguistic networks encode complex semantic hierarchies. Consequently, the structural bridgeheads in this regime are limited to highly invariant descriptors such as \textit{n\_nodes} and \textit{efficiency}, indicating that only the most abstract topological properties remain aligned between these two domains.

Across all directed pairs, a recurrent subset of descriptors consistently emerges as part of the shared structural backbone. In particular, \textit{efficiency}, $\lambda_2$, and \textit{modularity} appear robustly across domains, encoding common principles of information flow, spectral robustness, and mesoscale organization. Their repeated selection confirms that transferability is governed less by domain-specific motifs and more by shared constraints on how complex networks balance integration, resilience, and compartmentalization. Together, these results delineate the minimal set of interpretable structural anchors required for reliable manifold synchronization.

\subsection{Cross-Domain Knowledge Transfer and Performance Robustness}

\subsubsection{Cross-Domain Transfer Learning}

The evaluation of cross-domain knowledge propagation focuses exclusively on the parsimonious feature configuration (\textit{Top Feats}), isolating the impact of functionally grounded structural anchors from high-dimensional redundancy. All reported values represent ensemble averages marginalized across the full $10 \times 3 \times 3$ experimental grid, encompassing ten independent random seeds, data scarcity levels ($\alpha$), and stochastic corruption intensities ($\eta$).

At the global level, target-only models trained on aligned representations with structural anchors achieve near-optimal baseline performance (ROC-AUC $= 0.987$, AP $= 0.917$, F1-score $= 0.660$). Application of cross-domain transfer under the same configuration results in a moderate decrease in average metrics (ROC-AUC $= 0.974$, AP $= 0.879$, F1-score $= 0.559$) (Supplementary Table~\ref{tab:supp_regime_comparison}). This pattern identifies a pronounced ceiling effect: when separability in the target domain is already near-optimal, absolute improvements in mean performance are statistically constrained. Non-parametric evaluation validates the significance of the performance shifts observed across experimental scenarios. Kruskal–Wallis $H$-tests confirm significant variability for all metrics (ROC-AUC: $H = 188.44, p < 10^{-5}$; AP: $H = 125.06, p < 10^{-5}$; F1-score: $H = 202.93, p < 10^{-5}$).

Crucially, a comparative analysis between the full topological characterization (\textit{All Feats}) and the optimized structural anchors (\textit{Top Feats}) underscores the functional value of dimensionality control. While both configurations achieve near-optimal baseline performance in non-transfer scenarios, the structural anchors configuration demonstrates superior robustness under transfer-induced stress. Specifically, in transfer-learning tasks, the parsimonious selection of eight structural anchors achieves a higher global F1-score ($0.559$) compared to the full descriptor set ($0.556$). This performance gain confirms that the targeted exclusion of unstable discriminants prevents ``feature pollution'' during manifold synchronization. By filtering out domain-specific idiosyncrasies, the IIT strategy effectively regularizes the decision manifold, mitigating the F1-score collapse.

The observed discrepancy between ranking-based metrics (ROC-AUC $\approx 0.98$, AP $\approx 0.91$) and the threshold-dependent F1-score ($\approx 0.66$) is consistent with the inherent challenges of unsupervised anomaly detection under class imbalance. While the near-optimal ROC-AUC confirms the framework exceptional capacity to correctly rank anomalous instances based on shared structural anchors, the lower F1-score reflects the sensitivity of hard-decision boundaries to the 10\% contamination threshold (Sec.~\ref{subsec:anomaly_detection}). In transfer scenarios, the stability of ROC-AUC alongside the localized collapse of the F1-score identifies a ``threshold shift'' effect: while fundamental topological signatures of anomalies are preserved across disparate domains, the absolute scale of these descriptors undergoes non-linear transformations. Notably, the fact that the \textit{Top Feats} configuration maintains a higher F1-score ($0.559$) than the \textit{All Feats} set ($0.556$) reinforces the rescue effect of the IIT strategy, confirming that feature parsimony mitigates the impact of distributional noise on decision-making robustness. This stabilizing influence is significantly accentuated in regimes characterized by extreme data scarcity ($\alpha=0.1$) and high stochastic corruption ($\eta=0.9$), where the prioritization of domain-invariant structural anchors prevents the total collapse of the decision manifold that otherwise occurs when utilizing unrefined high-dimensional representations (Sec.~\ref{subsubsec:purity_noise}).

Analysis at the domain-pair level reveals localized positive gains and a high degree of coherence between the proposed scoring metrics and realized performance (Table~\ref{tab:comprehensive_transfer_results}).

\begin{table*}[h!]
\centering
\caption{\textbf{Comprehensive Cross-Domain Transfer Performance.} 
Ensemble-averaged transfer-learning results for 12 directed domain pairs using 8 structural anchors. 
Values report mean performance marginalized over 10 independent random seeds. 
NT denotes target-only training (No Transfer), whereas T denotes transfer learning from the source domain. 
Values in bold indicate a positive Transfer Gain Index (TGI) or transfer performance exceeding the corresponding NT baseline. 
Domain pairs follow the taxonomic order: Social (Soc), Molecular (Mol), Proteins (Prot), and Linguistic (Ling).}
\label{tab:comprehensive_transfer_results}

\scriptsize
\setlength{\tabcolsep}{1.8pt}
\renewcommand{\arraystretch}{1.2}

\begin{tabular}{l cc cc cc ccc}
\toprule
\textbf{Domain Pair} 
& \multicolumn{2}{c}{\textbf{ROC-AUC}} 
& \multicolumn{2}{c}{\textbf{AP}} 
& \multicolumn{2}{c}{\textbf{F1-score}} 
& \multicolumn{3}{c}{\textbf{Transfer Gain (TGI)}} \\
\cmidrule(lr){2-3} 
\cmidrule(lr){4-5} 
\cmidrule(lr){6-7} 
\cmidrule(lr){8-10}
(Src $\to$ Tgt) 
& \textbf{NT} & \textbf{T} 
& \textbf{NT} & \textbf{T} 
& \textbf{NT} & \textbf{T} 
& \textbf{ROC} & \textbf{F1} & \textbf{AP} \\
\midrule
Soc $\to$ Mol   
& 0.990 & 0.984 & 0.928 & 0.927 & 0.655 & 0.581 & -0.007 & -0.114 & -0.001 \\
Soc $\to$ Prot  
& 0.979 & 0.958 & 0.874 & 0.795 & 0.612 & 0.528 & -0.021 & -0.138 & -0.090 \\
Soc $\to$ Ling  
& 0.990 & 0.962 & 0.934 & 0.837 & 0.691 & 0.577 & -0.029 & -0.166 & -0.104 \\
\addlinespace
Mol $\to$ Soc   
& 0.989 & \textbf{0.993} & 0.928 & \textbf{0.950} & 0.678 & 0.529 & \textbf{0.005} & -0.220 & \textbf{0.023} \\
Mol $\to$ Prot  
& 0.985 & 0.947 & 0.898 & 0.819 & 0.617 & 0.492 & -0.038 & -0.202 & -0.088 \\
Mol $\to$ Ling  
& 0.983 & \textbf{0.985} & 0.899 & \textbf{0.906} & 0.683 & 0.554 & \textbf{0.002} & -0.189 & \textbf{0.007} \\
\addlinespace
Prot $\to$ Soc  
& 0.986 & 0.984 & 0.907 & \textbf{0.914} & 0.623 & 0.601 & -0.002 & -0.036 & \textbf{0.008} \\
Prot $\to$ Mol  
& 0.985 & 0.954 & 0.909 & 0.831 & 0.656 & 0.492 & -0.032 & -0.250 & -0.086 \\
Prot $\to$ Ling 
& 0.985 & 0.985 & 0.893 & \textbf{0.899} & 0.601 & \textbf{0.612} & 0.000 & \textbf{0.019} & \textbf{0.007} \\
\addlinespace
Ling $\to$ Soc  
& 0.992 & 0.992 & 0.946 & \textbf{0.950} & 0.701 & 0.596 & 0.000 & -0.150 & \textbf{0.004} \\
Ling $\to$ Mol  
& 0.983 & 0.961 & 0.886 & 0.830 & 0.635 & 0.542 & -0.023 & -0.146 & -0.064 \\
Ling $\to$ Prot 
& 0.991 & 0.982 & 0.941 & 0.898 & 0.709 & 0.627 & -0.009 & -0.117 & -0.046 \\
\bottomrule
\end{tabular}
\end{table*}

A high degree of coherence is observed between the $\text{IIT}_{\text{score}}$ rankings and the realized transfer utility. Notably, the \textit{Proteins $\to$ Linguistic} pair, which exhibited the highest structural affinity in the global consensus ($\overline{\text{IIT}}_{\text{score}} = 0.2470$), achieves synergistic gains (TGI$_{\text{F1}} = +0.019$, TGI$_{\text{AP}} = +0.007$). Similarly, transfers toward the \textit{Social} domain from \textit{Molecular} and \textit{Proteins} sources exhibit positive TGIs in ranking metrics (ROC-AUC and AP), confirming that the framework effectively regularizes dense social manifolds through uncorrupted structural priors. 

However, while the $\text{IIT}_{\text{score}}$ is inherently symmetric across domain pairs, empirical performance often exhibits pronounced directional asymmetry. This discrepancy identifies the source domain \textit{topological rigidity} as a decisive factor in transfer efficacy. Manifolds governed by stringent biophysical or chemical constraints, such as protein networks, act as high-fidelity scaffolds that effectively regularize target systems characterized by greater generative fluidity, such as social or linguistic networks. In contrast, transferring structural knowledge from more fluid domains back into high-constraint manifolds yields lower gains. This directional bias confirms that the \textit{rescue effect} is maximized when a stable, exogenous structural skeleton is available to anchor a corrupted target manifold, shifting the focus of transfer learning from mere similarity to the hierarchical stabilization of decision boundaries. The subsequent analysis (Sec.~\ref{subsubsec:tgi_iit_curve}) provides a granular investigation into these dynamics.

\subsubsection{Predictive Power of the IIT Score and Diversity-Driven Scaffolding}
\label{subsubsec:tgi_iit_curve}

\begin{figure}[H]
    \centering
    \includegraphics[width=0.85\textwidth]{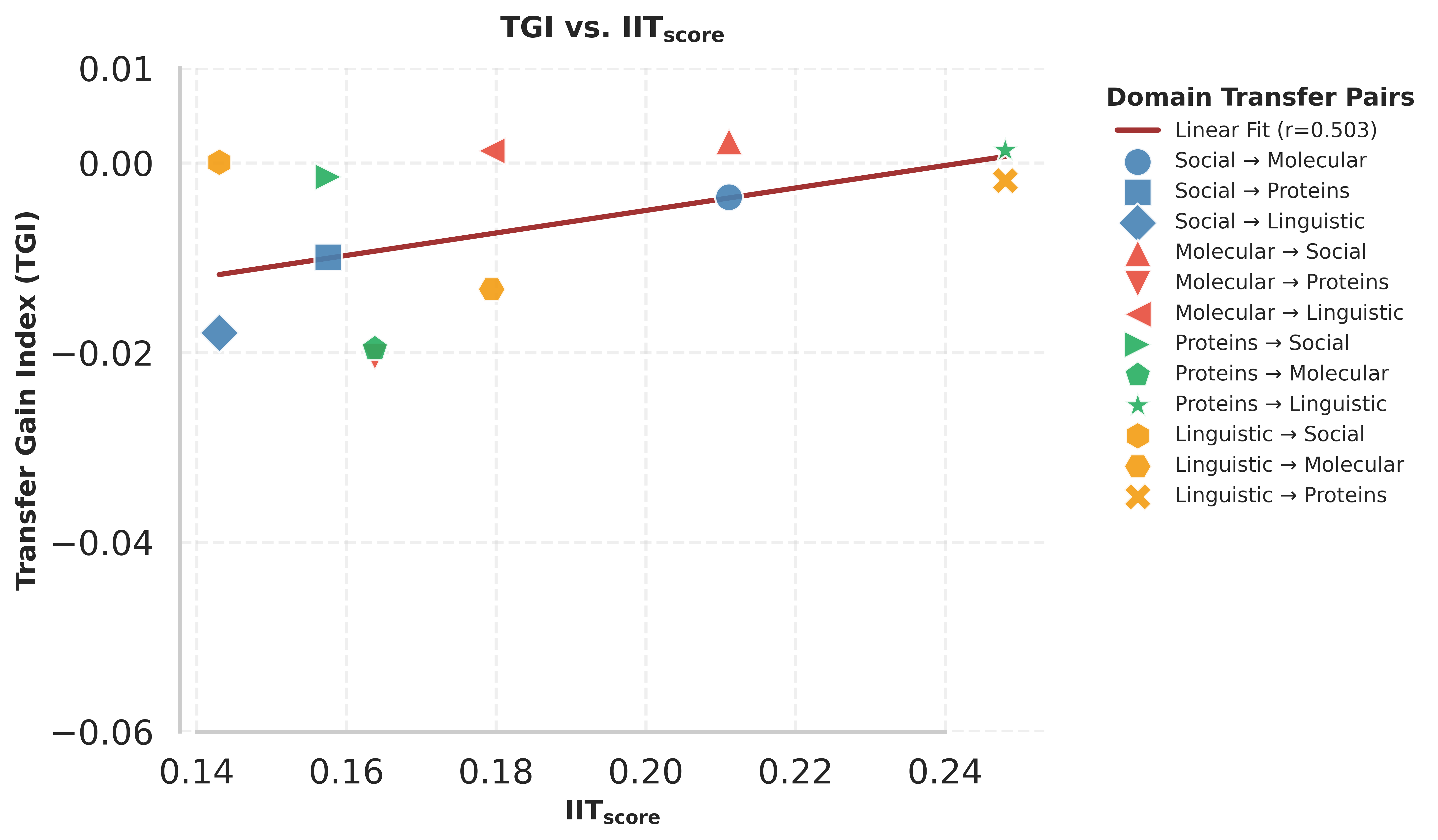}
    \caption{\textbf{Predictive validation of the IIT score.} Linear regression between the aggregate pairwise \textit{IIT Score} ($\overline{\text{IIT}}_{\text{score}}$) and the realized Transfer Gain Index (TGI). The substantial positive correlation ($r = 0.503$) identifies the $\text{IIT}_{\text{score}}$ as a robust predictor of transfer effectiveness. High-affinity pairs, notably \textit{Proteins $\to$ Linguistic} (green star), occupy the upper-right region, whereas outliers above the regression line identify a diversity-driven rescue effect where rigid structural skeletons regularize noisy target manifolds.}
    \label{fig:tgi_iit_regression}
\end{figure}

The association between theoretical manifold compatibility and empirical transfer effectiveness provides the most stringent validation of the proposed selection strategy. Figure~\ref{fig:tgi_iit_regression} reports the relationship between the aggregate pairwise \textit{IIT Score} ($\overline{\text{IIT}}_{\text{score}}$) and the realized \textit{Transfer Gain Index} (TGI), computed on ROC-AUC across the 12 directed domain pairs.

The regression analysis identifies a strong monotonic trend ($r = 0.503$), establishing the $\overline{\text{IIT}}_{\text{score}}$ as a good indicator of transfer performance. Despite a pronounced ceiling effect induced by the near-optimal performance of target-only models (ROC-AUC $\approx 0.98$), increasing $\text{IIT}_{\text{score}}$ values systematically mitigate transfer-induced degradation. This coherence is most evident in informational network pairs, such as \textit{Proteins $\leftrightarrow$ Linguistic}, which exhibit the highest aggregate structural bridge strength ($\overline{\text{IIT}}_{\text{score}} > 0.24$) and achieve positive synergistic gains. This finding confirms that informational networks share a latent hierarchical organization that facilitates robust distributed knowledge propagation.

A critical observation arises from the outliers situated above the regression line, specifically domain pairs that achieve high TGI despite moderate aggregate $\text{IIT}_{\text{score}}$. Most notably, the \textit{Molecular $\to$ Social} (red upward triangle) and \textit{Linguistic $\to$ Social} (orange hexagon) transfers exhibit performance gains that exceed expectations based purely on structural similarity. This phenomenon identifies a \emph{diversity-driven rescue effect}: the rigid, sparse, or hierarchical structural priors inherent in these source domains act as an exogenous topological scaffold. When a target social manifold is subjected to stochastic noise or data scarcity, these ``alien'' but internally consistent structures provide a robust discriminative grammar that prevents decision-boundary collapse. In these instances, the benefit of an uncorrupted structural skeleton outweighs the lack of metric proximity, confirming the \textit{transfer paradox} where diversity, rather than isomorphism, drives performance recovery under stress (cf. Sec.~\ref{subsubsec:purity_noise}).

Although the TGI is computed across the entire metric spectrum—including AP and F1-score—the current regression specifically utilizes the TGI derived from ROC-AUC. This choice is predicated on the threshold-independent nature of ROC-AUC, which directly reflects the geometry of the learned decision manifold rather than a specific operating point. Furthermore, ROC-AUC remains substantially more stable under the extreme data scarcity ($\alpha$) and noise ($\eta$) conditions investigated in the next section, whereas AP and F1-score are highly sensitive to threshold selection and positive class prevalence. Consequently, ROC-based TGI more faithfully captures the \textit{structural efficacy} of the transfer, serving as the most appropriate metric for linking theoretical compatibility to functional learning outcomes.

Overall, the positive correlation between the $\text{IIT}_{\text{score}}$ and TGI provides direct empirical support for the central hypothesis of X-CDTL: knowledge propagation is governed by the degree to which fundamental structural principles are shared across disparate domains. The $\text{IIT}_{\text{score}}$ thus emerges as a calibrated metric for evaluating structural homology, offering a quantitative bridge between explainable network organization and realized performance gains. 

\subsubsection{Operational Boundaries: High-Purity vs. High-Noise Regimes}
\label{subsubsec:purity_noise}

Evaluation of the X-CDTL framework across distinct operational regimes characterizes its functional boundaries. Analysis spans three macro-scenarios: the comprehensive experimental grid ($10 \times 3 \times 3$), a high-purity/data-scarce regime ($\eta=0.1, \alpha=0.1$), and a high-noise/high-data regime ($\eta=0.9, \alpha=0.9$) (Table~\ref{tab:global_boundary_comparison}).

\begin{table}[h!]
\centering
\caption{\textbf{Global Performance Comparison across Operational Regimes.} Marginalized averages for ROC-AUC, Average Precision (AP), and F1-score compare No Transfer (NT) and Transfer (T) scenarios using the optimized 8 structural anchors. Values represent the mean across 10 independent realizations. Bold values identify transfer performance exceeding the target-only baseline, highlighting the rescue effect in high-noise environments.}
\label{tab:global_boundary_comparison}
\small
\renewcommand{\arraystretch}{1.2}
\begin{tabular}{l cc cc cc}
\toprule
\textbf{Regime} & \multicolumn{2}{c}{\textbf{ROC-AUC}} & \multicolumn{2}{c}{\textbf{AP}} & \multicolumn{2}{c}{\textbf{F1-score}} \\
\cmidrule(lr){2-3} \cmidrule(lr){4-5} \cmidrule(lr){6-7}
& \textbf{NT} & \textbf{T} & \textbf{NT} & \textbf{T} & \textbf{NT} & \textbf{T} \\
\midrule
Full Experimental Grid              & 0.987 & 0.974 & 0.917 & 0.879 & 0.660 & 0.559 \\
High-Purity ($\eta=0.1, \alpha=0.1$) & 0.983 & 0.953 & 0.896 & 0.819 & 0.760 & 0.312 \\
High-Noise ($\eta=0.9, \alpha=0.9$)  & 0.990 & 0.985 & 0.933 & 0.916 & 0.341 & \textbf{0.532} \\
\bottomrule
\end{tabular}
\end{table}

\paragraph{High-Purity Regime and Distributional Interference.}
In environments characterized by high signal purity and minimal training samples ($N=50$), the results identify a transition from the structural regularization observed under stress to a regime of \textit{distributional interference}. In this uncorrupted setting, ``No Transfer'' scenarios achieve near-optimal accuracy (ROC-AUC $\approx 0.98$, F1 $= 0.760$), indicating that the intrinsic structural integrity of the target ensemble is sufficient for precise anomaly detection without external knowledge.
Conversely, the integration of 500 source-domain samples induces significant \textit{negative transfer} in threshold-dependent metrics, with the global F1-score collapsing to $0.312$. This decay suggests that the source-to-target sample imbalance ($10:1$) causes the source domain structural mass to overwhelm the target local signal, acting as a ``topological blur'' on the decision manifold. Non-parametric inference confirms the significance of these performance shifts across all metrics (Kruskal-Wallis ROC-AUC: $H=133.31$; AP: $H=108.18$; F1: $H = 292.52, p < 0.0001$). 

Despite the performance decay in absolute terms, regression analysis reveals that theoretical manifold compatibility continues to govern transfer stability. The association between the aggregate pairwise IIT$_{\text{score}}$ and the TGI remains positive ($r = 0.372$; Supplementary Fig.\ref{fig:supp_reg_low_noise}), indicating that higher predicted affinity mitigates the intensity of negative transfer even in high-purity regimes. 
This coherence is most evident in the \textit{Proteins $\to$ Linguistic} pair, which maintains the highest directed $\text{IIT}_{\text{score}}$ ($0.248$). In this specific case, the framework achieves synergistic gains despite the high-purity baseline, with the F1-score increasing from $0.743$ to $0.809$ ($\text{TGI}_{\text{F1}} = +0.089$) and Average Precision rising from $0.895$ to $0.913$ (Table~\ref{tab:comprehensive_transfer_results_purity}). 

\begin{table*}[h!]
\centering
\caption{\textbf{Comprehensive Cross-Domain Transfer Performance in High-Purity Regime ($\eta=0.1, \alpha=0.1$).} 
Ensemble-averaged results for 12 directed domain pairs using the optimized 8 structural anchors. 
NT denotes target-only training, whereas T denotes transfer learning. 
Values in bold indicate a positive Transfer Gain Index (TGI). 
Taxonomic sequence: Social (Soc), Molecular (Mol), Proteins (Prot), and Linguistic (Ling).}
\label{tab:comprehensive_transfer_results_purity}
\scriptsize
\setlength{\tabcolsep}{1.8pt}
\renewcommand{\arraystretch}{1.2}
\begin{tabular}{l cc cc cc ccc}
\toprule
\textbf{Domain Pair} & \multicolumn{2}{c}{\textbf{ROC-AUC}} & \multicolumn{2}{c}{\textbf{AP}} & \multicolumn{2}{c}{\textbf{F1-score}} & \multicolumn{3}{c}{\textbf{Transfer Gain (TGI)}} \\
\cmidrule(lr){2-3} \cmidrule(lr){4-5} \cmidrule(lr){6-7} \cmidrule(lr){8-10}
(Src $\to$ Tgt) & \textbf{NT} & \textbf{T} & \textbf{NT} & \textbf{T} & \textbf{NT} & \textbf{T} & \textbf{ROC} & \textbf{F1} & \textbf{AP} \\
\midrule
Soc $\to$ Mol   & 0.988 & 0.966 & 0.917 & 0.882 & 0.798 & 0.357 & -0.022 & -0.553 & -0.038 \\
Soc $\to$ Prot  & 0.973 & 0.932 & 0.859 & 0.716 & 0.713 & 0.213 & -0.042 & -0.702 & -0.167 \\
Soc $\to$ Ling  & 0.991 & 0.931 & 0.937 & 0.759 & 0.757 & 0.281 & -0.060 & -0.629 & -0.190 \\
\addlinespace
Mol $\to$ Soc   & 0.978 & \textbf{0.990} & 0.873 & \textbf{0.921} & 0.767 & 0.186 & \textbf{0.012} & -0.757 & \textbf{0.055} \\
Mol $\to$ Prot  & 0.986 & 0.883 & 0.908 & 0.684 & 0.781 & 0.192 & -0.105 & -0.754 & -0.247 \\
Mol $\to$ Ling  & 0.980 & \textbf{0.981} & 0.889 & 0.885 & 0.768 & 0.207 & \textbf{0.001} & -0.731 & -0.004 \\
\addlinespace
Prot $\to$ Soc  & 0.979 & 0.964 & 0.865 & 0.827 & 0.738 & 0.250 & -0.015 & -0.661 & -0.044 \\
Prot $\to$ Mol  & 0.986 & 0.919 & 0.910 & 0.762 & 0.766 & 0.412 & -0.068 & -0.461 & -0.163 \\
Prot $\to$ Ling & 0.983 & \textbf{0.986} & 0.895 & \textbf{0.913} & 0.743 & \textbf{0.809} & \textbf{0.003} & \textbf{0.089} & \textbf{0.021} \\
\addlinespace
Ling $\to$ Soc  & 0.987 & \textbf{0.988} & 0.917 & \textbf{0.920} & 0.805 & 0.226 & \textbf{0.001} & -0.720 & \textbf{0.003} \\
Ling $\to$ Mol  & 0.981 & 0.938 & 0.866 & 0.756 & 0.718 & 0.318 & -0.043 & -0.557 & -0.128 \\
Ling $\to$ Prot & 0.988 & 0.956 & 0.920 & 0.807 & 0.767 & 0.294 & -0.033 & -0.617 & -0.124 \\
\bottomrule
\end{tabular}
\end{table*} 

Such findings suggest that the shared hierarchical organization identified by the $\text{IIT}_{\text{score}}$ creates a resilient topological bridge that resists distributional interference. 
Furthermore, a unique ``rescue effect'' in ranking capacity is observed in the \textit{Molecular $\to$ Social} and \textit{Linguistic $\to$ Social} transfers. For these pairs, the integration of source knowledge leads to positive gains in ROC-AUC ($\text{TGI}_{\text{ROC}} = +0.012$ and $+0.001$, respectively) and Average Precision. This indicates that while the absolute decision threshold (F1) is compromised by the sample imbalance, the underlying structural grammar provided by the source domain improves the model ability to correctly rank anomalous instances in the target social manifold.

\paragraph{High-Noise Regime and the Structural Rescue Effect.}
Performance dynamics under maximum feature corruption ($\eta=0.9, \alpha=0.9$) reveal a fundamental decoupling between ranking capacity and classification robustness. While the isolated target baseline maintains a near-optimal ROC-AUC ($0.990$) and Average Precision ($0.933$), the F1-score collapses to $0.341$ (Table~\ref{tab:global_boundary_comparison}). This divergence indicates that pervasive noise prevents the establishment of stable decision thresholds despite high underlying class separability. 
The integration of source knowledge through the IIT logic triggers a dramatic recovery, with the global F1-score increasing from $0.341$ to $0.532$ when utilizing the optimized structural anchors. Non-parametric inference confirms the high statistical significance of this recovery for the F1 metric ($H = 83.80, p < 0.0001$), while shifts in ROC-AUC ($H = 7.62, p = 0.055$) and AP ($H = 6.68, p = 0.083$) remain statistically non-significant. These results validate that the primary utility of knowledge transfer in high-noise environments lies in regularizing the decision manifold rather than improving already saturated ranking metrics.

Detailed mapping across the 12 directed domain pairs (Table~\ref{tab:comprehensive_transfer_results_noise}) identifies specific topological bridgeheads where the source acts as a stabilizing scaffold. 
\begin{table*}[h!]
\centering
\caption{\textbf{Comprehensive Cross-Domain Transfer Performance in High-Noise Regime ($\eta=0.9, \alpha=0.9$).} 
Ensemble-averaged transfer-learning results for 12 directed domain pairs using 8 structural anchors. 
NT denotes target-only training, whereas T denotes transfer learning. 
Values in bold indicate a positive Transfer Gain Index (TGI). 
Taxonomic sequence: Social (Soc), Molecular (Mol), Proteins (Prot), and Linguistic (Ling).}
\label{tab:comprehensive_transfer_results_noise}
\scriptsize
\setlength{\tabcolsep}{1.8pt}
\renewcommand{\arraystretch}{1.2}
\begin{tabular}{l cc cc cc ccc}
\toprule
\textbf{Domain Pair} 
& \multicolumn{2}{c}{\textbf{ROC-AUC}} 
& \multicolumn{2}{c}{\textbf{AP}} 
& \multicolumn{2}{c}{\textbf{F1-score}} 
& \multicolumn{3}{c}{\textbf{Transfer Gain (TGI)}} \\
\cmidrule(lr){2-3} 
\cmidrule(lr){4-5} 
\cmidrule(lr){6-7} 
\cmidrule(lr){8-10}
(Src $\to$ Tgt) 
& \textbf{NT} & \textbf{T} 
& \textbf{NT} & \textbf{T} 
& \textbf{NT} & \textbf{T} 
& \textbf{ROC} & \textbf{F1} & \textbf{AP} \\
\midrule
Soc $\to$ Mol   & 0.992 & 0.992 & 0.928 & \textbf{0.945} & 0.305 & \textbf{0.329} & 0.000 & \textbf{0.078} & \textbf{0.019} \\
Soc $\to$ Prot  & 0.975 & \textbf{0.979} & 0.877 & 0.860 & 0.323 & \textbf{0.546} & \textbf{0.004} & \textbf{0.694} & -0.020 \\
Soc $\to$ Ling  & 0.996 & 0.975 & 0.971 & 0.873 & 0.488 & \textbf{0.595} & -0.021 & \textbf{0.220} & -0.101 \\
\addlinespace
Mol $\to$ Soc   & 0.996 & \textbf{0.998} & 0.963 & \textbf{0.981} & 0.415 & \textbf{0.736} & \textbf{0.002} & \textbf{0.772} & \textbf{0.019} \\
Mol $\to$ Prot  & 0.978 & \textbf{0.985} & 0.858 & \textbf{0.922} & 0.224 & \textbf{0.603} & \textbf{0.007} & \textbf{1.695} & \textbf{0.074} \\
Mol $\to$ Ling  & 0.992 & 0.991 & 0.947 & 0.941 & 0.395 & \textbf{0.674} & -0.001 & \textbf{0.705} & -0.007 \\
\addlinespace
Prot $\to$ Soc  & 0.991 & \textbf{0.992} & 0.941 & \textbf{0.948} & 0.305 & \textbf{0.635} & \textbf{0.001} & \textbf{1.083} & \textbf{0.007} \\
Prot $\to$ Mol  & 0.991 & 0.973 & 0.933 & 0.853 & 0.370 & 0.190 & -0.017 & -0.485 & -0.086 \\
Prot $\to$ Ling & 0.986 & 0.986 & 0.901 & 0.899 & 0.289 & 0.289 & 0.000 & 0.000 & -0.003 \\
\addlinespace
Ling $\to$ Soc  & 0.998 & 0.996 & 0.988 & 0.971 & 0.370 & \textbf{0.795} & -0.002 & \textbf{1.149} & -0.017 \\
Ling $\to$ Mol  & 0.987 & 0.973 & 0.914 & 0.875 & 0.271 & \textbf{0.346} & -0.014 & \textbf{0.276} & -0.043 \\
Ling $\to$ Prot & 0.995 & 0.984 & 0.968 & 0.922 & 0.332 & \textbf{0.650} & -0.010 & \textbf{0.956} & -0.047 \\
\bottomrule
\end{tabular}
\end{table*}
Notably, the most substantial improvements occur in transfers originating from the \textit{Molecular} and \textit{Linguistic} domains. For instance, the \textit{Molecular $\to$ Proteins} transfer exhibits a massive F1-score gain ($\text{TGI}_{\text{F1}} = +1.695$), while the \textit{Linguistic $\to$ Social} and \textit{Proteins $\to$ Social} transfers achieve F1 gains of $+1.149$ and $+1.083$, respectively. 

Regression analysis confirms a positive correlation between theoretical manifold compatibility and empirical transfer gains ($r = 0.135$; Supplementary Fig.~\ref{fig:supp_reg_high_noise}). This trend suggests that even in extreme noise regimes, knowledge propagation effectiveness is governed by the structural bridge strength identified via the $\text{IIT}_{\text{score}}$. In this scenario, an externally consistent source manifold provides a rigid ``structural skeleton'' that compensates for the noise-induced collapse of local decision boundaries. This exogenous prior imposes a stable structural grammar that pervasive target noise cannot obscure, proving that the identification of domain-invariant anchors is the primary driver of manifold regularization under stress.

\section{Discussion}
\label{sec:discussion}

Benchmark analyses provide evidence for a robust, cross-disciplinary topological signature inherent in complex systems. The high consistency of ROC-AUC scores---exceeding $0.97$ in both target-only and transfer scenarios---demonstrates that the selected feature set captures an intrinsic fingerprint of network organization. The minimal performance degradation during transfer indicates that while specific decision thresholds vary across manifolds, the underlying structural signal remains exceptionally stable across disciplines. Such effectiveness arises from the IIT mechanism, which identifies topological descriptors agnostic to the domain of origin. These structural anchors act as shared organizational constants that mitigate feature pollution, enabling the model to bypass domain-specific idiosyncrasies in favor of a shared architectural backbone. This finding aligns with the theory of self-organized criticality; as established by Jensen~\cite{jensen1998self}, complex systems naturally evolve toward universal dynamical states, resulting in substrate-independent organizational principles that the X-CDTL framework identifies as stable topological invariants.

The integration of XML into these transfer learning workflows enables two distinct strategies for knowledge propagation. The first, formalized in \cite{caligiore2025exploring}, relies on \textit{phenomenological alignment} by identifying the \emph{most important} topological features characterizing a shared process, such as criticality. In contrast, the current work introduces a \textit{structural invariance} paradigm through the IIT logic. This strategy employs an inverted selection principle: rather than prioritizing features that best characterize a specific state, the framework isolates descriptors with the \emph{lowest} discriminative utility for domain classification. Features that fail to distinguish between different network classes effectively encode stable structural anchors. By filtering out highly discriminative features in favor of these constants, the IIT mechanism establishes a robust scaffold for transfer between radically heterogeneous domains. This approach positions X-CDTL within the broader category of feature-representation-based transfer learning~\cite{pan2009survey}, yet the use of explainability as an active selection mechanism transforms abstract representations into interpretable mathematical foundations.

The framework operates across two hierarchical levels of structural identification. Globally, the \textit{Global Consensus} $\text{IIT}_{\text{score, G}}$ identifies the shared topological backbone---the anchors that remain invariant across all investigated generative dynamics simultaneously (Fig.~\ref{fig:global_ranking}). Locally, the pair-specific $\text{IIT}_{\text{score}}$ refines this selection for specific source--target transitions, optimizing manifold alignment by accounting for unique distributional overlaps. A critical technical distinction exists here between topological invariance and numerical synchronization. While topological consistency is maintained, invariant properties often manifest across divergent scales. The dual-stage pipeline addresses this through GPU-accelerated PCA-SVD mapping, which synchronizes the numerical ``voltages'' of source and target distributions. This synchronization neutralizes the distributional drag arising from parameter-scale mismatches, ensuring that knowledge transfer remains theoretically grounded and numerically stable.

The interpretation of the TGI requires a nuanced understanding of domain-specific baselines. Negative TGI values do not indicate a failure of knowledge propagation. Instead, they reflect a \textit{ceiling effect} where the intrinsic structural integrity of the target domain is already near-optimal (ROC-AUC $\approx 0.98$). In these uncorrupted settings, the success of X-CDTL is evidenced by the conservation of ranking stability (ROC-AUC $> 0.95$), proving that the identified structural anchors provide a common architectural backbone that remains functional even across fundamentally different generative dynamics. In this sense, the IIT mechanism acts as a structural regularizer: it stabilizes the decision manifold by providing a consistent topological prior that prevents the model from overfitting to idiosyncratic local features.

The emergence of the \textit{transfer paradox}---where optimal generalization occurs at intermediate structural similarity---delineates a critical boundary between topological redundancy and functional novelty. Regression analysis reveals a sophisticated rescue effect governed by this balance. In high-purity regimes, a positive correlation between the directed $\text{IIT}_{\text{score}}$ and Transfer Gain ($r = 0.372$) suggests that similarity acts as a buffer against negative transfer, preventing domain-specific noise from degrading target performance~\cite{weiss2016survey}. 

Conversely, under maximum feature corruption ($\eta = 0.9$), the relationship remains positive and identifies a significant \emph{Diversity-Driven Rescue Effect}. In extreme noise scenarios, the target domain undergoes a ``topological blur'' that obscures local feature-level similarity. Here, a distant but internally consistent source manifold provides a rigid, uncorrupted ``topological prosthesis.'' Because these structures remain independent of local corruption, they impose a stable structural grammar that pervasive noise cannot obscure. 
This performance recovery is most pronounced in high-noise regimes, where the global F1-score increases from $0.341$ to $0.532$---representing a \emph{56\% relative improvement} over target-only benchmarks. This substantial gain in F1-score, contrasted with the relative stability of ROC-AUC, identifies \textit{decision stabilization} as a fundamental contribution of the X-CDTL framework. In these corrupted environments, isolated target models experience a characteristic \emph{threshold collapse}: while they maintain the capacity to correctly rank anomalous instances (as evidenced by high ROC-AUC), they fail to establish a stable decision boundary. By providing an uncorrupted structural skeleton from the source domain, the IIT mechanism regularizes the target manifold. This exogenous prior effectively rescues the decision threshold, preserving the model functional integrity even under maximum stochastic corruption.

\subsection{Topological Scaffolding and Cross-Domain Horizons: From Social-Molecular Synergies to the Grammar of Proteins}

Beyond theoretical discovery, the X-CDTL framework provides an analytical scaffold for scientific fields where high-fidelity labeled data is economically or ethically constrained~\cite{caccia2023towards}. The substantial rescue effect confirms that common structural anchors can regularize target manifolds even when local signals are nearly destroyed. In the \emph{pharmaceutical sciences}, this suggests that dense \emph{social ego-networks} could serve as topological scaffolds for \emph{molecular graphs} in drug-discovery tasks. By utilizing the triadic redundancy of social structures to regularize sparse molecular connectivity, X-CDTL could facilitate the identification of bioactive candidates in under-determined chemical spaces. Similarly, the high structural affinity identified between \emph{protein interaction networks} and \emph{linguistic co-occurrence networks} ($\overline{\text{IIT}}_{\text{score}} = 0.248$) suggests that the ``syntactic grammar'' of human language could be repurposed to predict protein-protein contacts~\cite{rives2021biological}, acting as a regularizing prior for high-throughput contact data.

The utility of this paradigm extends to \emph{public health engineering}, where the rigid, acyclic constraints of \emph{molecular topologies} can impose a stable discriminative logic on sparse or corrupted \emph{social-network data}. This could improve epidemiological diffusion models by anchoring fluctuating social dynamics to the stable connectivity laws of physical systems. Furthermore, the capacity to isolate domain-invariant structural anchors positions the X-CDTL framework as a powerful diagnostic instrument for identifying common mechanistic pathways underlying disparate disease states in \emph{clinical neuroscience}. For example, the IIT mechanism is uniquely suited to testing the \textit{Neurodegenerative Elderly Syndrome} (NES) hypothesis, which posits that Alzheimer’s and Parkinson’s diseases represent divergent manifestations of a unified pathological process~\cite{caligiore2022neurodegenerative}. By prioritizing domain-invariant structural anchors, the framework provides a methodology to uncover the shared topological signature of brain network degradation~\cite{barabasi2011network,ghiassian2015network}, potentially identifying unified biomarkers that remain obscured by traditional models focusing on phenotypic differences.

\section{Conclusion}
\label{sec:conclusion}

The development of the X-CDTL framework demonstrates that common organizational principles transcend disciplinary boundaries, enabling robust knowledge propagation through the IIT mechanism. By isolating a parsimonious set of domain-invariant structural anchors, the framework achieves a 56\% relative improvement in decision stability under extreme noise. This confirms that interpretable topological constants provide a more resilient foundation for cross-domain generalization than traditional opaque embeddings, effectively bridging disparate generative dynamics---from molecular branching to social triadic closure.

While the current implementation addresses static, undirected topologies, it establishes a versatile basis for future research into dynamic, multilayer, and causal architectures. Subsequent efforts will target the synthesis of this domain-invariant framework with mechanism-centric transfer strategies \cite{caligiore2025exploring}, aiming to further enhance the mechanistic interpretability of general organizational principles inherent in complex networked systems.

\section{Methods}
\subsection{Overview of the X-CDTL Framework}

\begin{figure}[ht!]
    \centering \includegraphics[width=\textwidth]{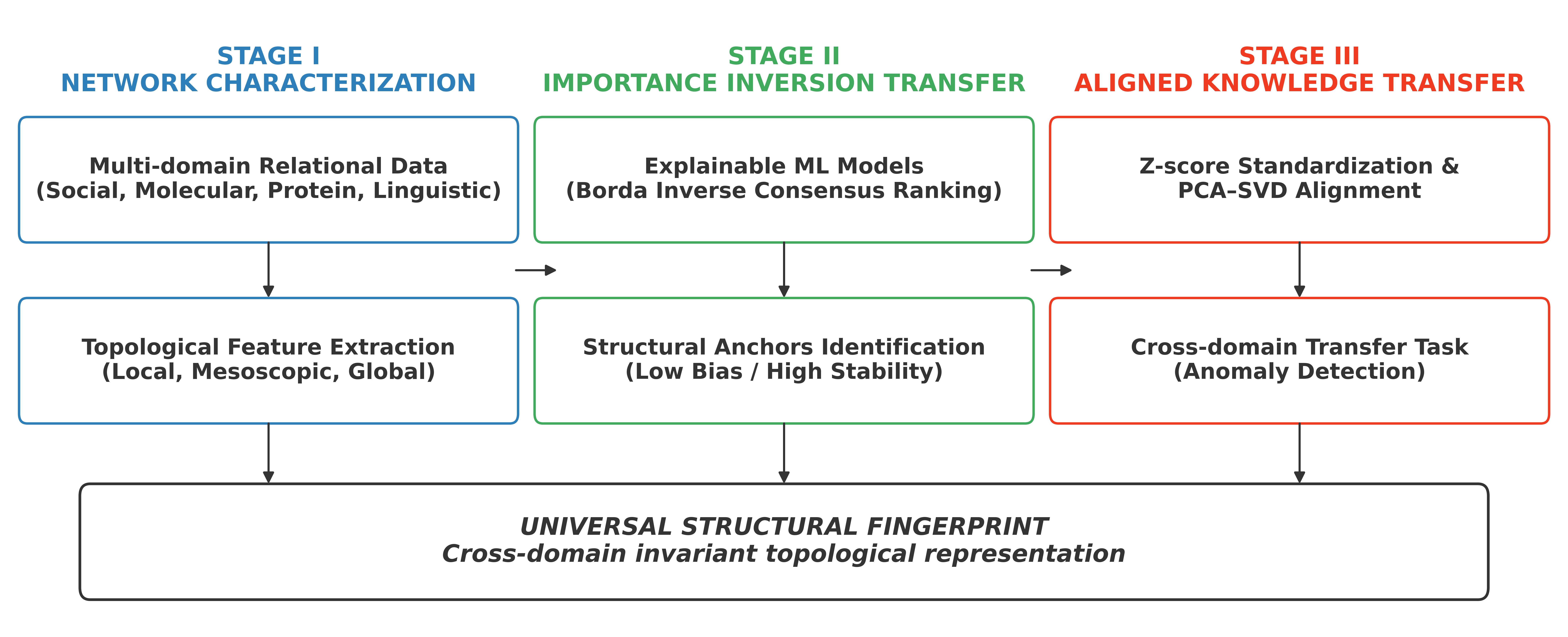}
    \caption{\textbf{Workflow of the IIT strategy within the X-CDTL framework.} 
    The pipeline is organized into three sequential modules. 
    Stage I: Network Characterization. Relational data from social, molecular, protein, and linguistic domains are transformed into a unified graph-based representation to extract multi-scale topological descriptors spanning local, mesoscopic, and global scales. 
    Stage II: Importance Inversion Transfer (IIT). A consensus-based Borda \textit{Inverse} Ranking strategy is employed to identify structural anchors by prioritizing features with the lowest discriminative utility for domain classification and the highest cross-domain stability. 
    Stage III: Aligned Knowledge Transfer. Target and source manifolds undergo z-score standardization and PCA-SVD alignment. This manifold synchronization enables robust cross-domain anomaly detection under severe data scarcity and noise, effectively yielding a common structural fingerprint (structural anchors) that captures shared organizational principles across disparate scientific fields.}
    \label{fig:pipeline_xcdtl}
\end{figure}

Explainable Cross-Domain Transfer Learning (X-CDTL) provides a principled route for interpretable knowledge transfer across heterogeneous scientific domains operating under limited or noisy data regimes.  
The framework transforms complex systems into comparable graph representations, identifies invariant structural features, and quantifies how these shared principles enable transfer across disciplinary boundaries.  
The workflow unfolds through three integrated modules (Fig.~\ref{fig:pipeline_xcdtl}):  
(i) topological characterization of domain-specific systems,  
(ii) identification of stable and transferable structural anchors, and (iii) domain-aligned transfer learning guided by interpretable graph topology.  

This architecture emphasizes explicit structural knowledge rather than latent embeddings, ensuring that generalization arises from measurable topological organization rather than opaque feature transformations.  
The design enables functional interpretability and quantitative assessment of transferability, forming a bridge between explainable artificial intelligence and theoretical network science.

\subsection{Domains and Network Construction}

Four representative domains were selected to span diverse scientific scales and generative mechanisms: social and molecular graphs, protein--protein interaction networks, and linguistic co-occurrence networks. Each system is represented as a complex network $G=(V,E)$, allowing a unified structural abstraction that isolates topological organisation from domain-specific semantics or physical meaning. In this notation, $V$ denotes the set of fundamental entities---such as users, atoms, amino acids, or words---while $E$ represents the set of relational interactions, including social ties, chemical bonds, physical contacts, or contextual co-occurrences, respectively.

\paragraph{Social networks.} 
Human relational dynamics are modeled using ego-centric subgraphs extracted from the Facebook dataset\footnote{\url{https://snap.stanford.edu/data/ego-Facebook.html}} provided by the Stanford Network Analysis Project (SNAP) \cite{leskovec2012learning}. In this domain, nodes represent individual users and edges denote mutual friendship ties. These networks are characterized by intense local redundancy and high clustering coefficients, arising primarily from triadic closure mechanisms and preferential attachment \cite{barabasi1999emergence}. By sampling ego-networks within a controlled size range (10--30 nodes), the framework captures the mesoscopic organization of social ``circles,'' which exhibit small-world properties and a quasi-clique architecture that stands in stark contrast to valence-constrained physical systems.

\paragraph{Molecular networks.} 
Chemical structures are sourced from the QM9 database\footnote{\url{https://moleculenet.org/datasets-1}}, which comprises approximately 134,000 small organic molecules characterized by up to nine heavy atoms (C, O, N, and F) \cite{ramakrishnan2014qm9}. Graphs are constructed with nodes representing atoms and edges representing covalent chemical bonds. These structures provide low-noise, physically constrained exemplars of topology where connectivity is strictly governed by quantum mechanical valence rules and spatial embedding. The resulting acyclic or sparingly cyclic configurations offer a benchmark for sparse, highly constrained branching patterns.

\paragraph{Protein--protein interaction networks.} 
Biological signaling and structural motifs are represented using the PROTEINS dataset\footnote{\url{https://chrsmrrs.github.io/datasets/}} from the TUDataset repository \cite{morris2020tudataset, borgwardt2005protein}. In these graphs, nodes represent secondary structure elements, and edges denote physical proximity or functional interactions between them. These networks exhibit a distinctive modular organization and hierarchical topology reflective of biological functionality and evolutionary optimization. The presence of well-defined functional domains allows for the evaluation of transferability based on mesoscale compartmentalization.

\paragraph{Linguistic co-occurrence networks.} 
Syntactic and semantic structures are derived from the \textit{Standard Corpus of Present-Day Edited American English} (the Brown Corpus), a seminal dataset for the computational frequency analysis of English usage \cite{kontra1985review}. Accessed via the Natural Language Toolkit (NLTK) distribution\footnote{\url{https://www.nltk.org/data.html}}, the corpus represents a balanced collection of diversified prose styles. This dataset enables the construction of word-adjacency graphs where nodes represent unique lexical tokens and edges denote contextual co-occurrence within a fixed sliding window. These graphs capture the relational topology of human language, reflecting syntactic dependencies and communicative efficiency without the constraints of manual annotation. The resulting topologies provide a rigorous benchmark for informational networks shaped by linguistic hierarchy and scale-free organization.

\addvspace{10pt}
\noindent All graphs are treated as undirected and unweighted to ensure cross-domain comparability. Only structural features are retained, ensuring that knowledge propagation relies exclusively on topological invariants rather than domain-specific semantic labels or node attributes.

\begin{table}[h!]
\centering
\caption{\textbf{Overview of datasets and graph characteristics across domains.} Values indicate the typical scale and structural heterogeneity of each system. The total number of graphs reflects the ensemble size aggregated over 10 independent stochastic realizations ($N=500$ per seed).}
\label{tab:datasets}
\small
\renewcommand{\arraystretch}{1.2}
\begin{tabular}{lcccc}
\toprule
\textbf{Domain} & \textbf{Dataset} & \textbf{Nodes/Graph} & \textbf{Edges/Graph} & \textbf{Graphs} \\
\midrule
Social & Facebook Ego & 10--30 & 19--404 & 5,000 \\
Molecular & QM9 & 10--23 & 9--22 & 5,000 \\
Proteins & PROTEINS & 10--30 & 12--72 & 5,000 \\
Linguistic & Brown Corpus & 10--30 & 16--72 & 5,000 \\
\bottomrule
\end{tabular}
\end{table}

\subsection{Network Feature Extraction}

The structural characterization of each network ensemble relies on the quantification of twelve topological descriptors spanning local, mesoscopic, and global scales. This multi-scale approach projects each system into a unified structural manifold, facilitating direct comparison across disparate scientific domains. The feature set comprises:

\begin{itemize}
    \item[] \emph{Scale and Connectivity:} Node count ($|V|$), edge count ($|E|$), and graph density define the fundamental scale and sparsity of the topology.
    \item[] \emph{Segregation and Local Cohesion:} The average clustering coefficient and network transitivity quantify the prevalence of localized triangles and the tendency for triadic closure.
    \item[] \emph{Degree-Mixing Patterns:} The assortativity coefficient characterizes the preference of nodes to connect with others of similar degree, reflecting the network degree-degree correlation structure.
    \item[] \emph{Integrative Capacity and Path Metrics:} Global efficiency, mean shortest path length, and network diameter describe the efficiency of information propagation and the global integration of the manifold.
    \item[] \emph{Spectral Fingerprints:} The spectral radius (the largest eigenvalue of the adjacency matrix) and algebraic connectivity ($\lambda_2$ of the Laplacian matrix) provide insights into the network robustness, diffusion properties, and synchronization potential.
    \item[] \emph{Mesoscopic Organization:} The modularity index, derived via the Louvain partitioning algorithm, identifies the strength of hierarchical community division within the graph.
\end{itemize}

Standardization via $z$-score normalization ensures numerical comparability across heterogeneous graph domains by mitigating scale discrepancies. Furthermore, median imputation maintains data integrity for non-computable metrics in specific topological configurations (e.g., disconnected components), providing a numerically stable input for the subsequent transfer learning pipeline. 

\subsection{Feature Selection, Stability, and Transferability Analysis}

\subsubsection{Feature Selection and Topological Parsimony}

The identification of structural anchors relies on a strategic balance between topological expressivity and statistical robustness. Iterative benchmarking of twenty graph-theoretical metrics—encompassing spectral, path-based, and connectivity properties—reveals that expanding the feature space beyond twelve primary descriptors consistently degrades performance, particularly in cross-domain settings. This instability arises from two primary mechanisms: domain-specific scale noise and multicollinearity. Magnitude disparities in high-variance descriptors, such as the Wiener Index, introduce scale noise that biases decision boundaries, while feature redundancy facilitates overfitting to domain-specific artifacts (cf., Sec.~\ref{subsubsec:dimensionality_control}).

To mitigate these limitations, the framework implements a multi-stage selection pipeline centered on the $\text{IIT}_{\text{score}}$ (Sec.~\ref{subsubsec:steps}) to isolate eight structural anchors from the primary twelve-metric pool. This 12/8 configuration represents a \textit{topological sweet spot}: a parsimonious subspace that maximizes cross-domain generalizability while insulating the model from scale-induced noise. Consequently, downstream models operate within a structurally stable representation optimized for robust knowledge propagation across heterogeneous scientific datasets. 

\subsubsection{Domain-Aware Feature Selection and Manifold Alignment}
\label{subsubsec:steps}

\paragraph{Conceptual Premise of Importance Inversion.} 
The IIT mechanism is based on the following heuristic principle: if a topological descriptor $f_i$ exhibits high discriminative utility for domain classification, it effectively encodes domain-specific idiosyncrasies that are unlikely to generalize. Conversely, descriptors that fail to distinguish between disparate domains—while remaining structurally informative—identify the \textit{domain-neutral} properties. By prioritizing features with the lowest discriminative utility, the framework isolates the structural invariants that constitute a stable backbone for knowledge transfer.

The identification of transferable structural anchors proceeds through a hierarchical evaluation of topological relevance, statistical stability, and manifold compatibility. This multi-stage pipeline ensures that the resulting feature subset maximizes predictive utility while maintaining structural invariance across disparate generative dynamics.

\paragraph{Step 1: Quantifying global relevance via Borda aggregation.}
To identify descriptors with varying degrees of discriminative utility across the domain landscape, a consensus-based importance score, $B_i$, is calculated. Three supervised classifiers—Random Forest (RF), Gradient Boosting (GB), and Logistic Regression (LR)—are trained to predict the domain of origin based exclusively on topological attributes. These specific models are selected to provide a diversity of inductive biases: RF and GB capture complex, non-linear feature interactions through bagging and boosting ensembles, respectively, while LR ensures that fundamental linear separability is accounted for. This ensemble-based approach mitigates model-specific artifacts, ensuring a robust estimation of feature relevance. 

The $B_i$ score for each descriptor $f_i$ is defined by its average importance rank across these $M=3$ architectures:
\begin{equation}
    B_i = \frac{1}{M} \sum_{m=1}^{M} r_{i,m},
\end{equation}
where $r_{i,m}$ denotes the importance rank of feature $f_i$ in model $m$ (e.g., $r=1$ for the most discriminative feature and $r=12$ for the least). Crucially, following the logic of Importance Inversion Transfer (IIT), the Borda score is utilized in an \textit{inverse} manner. While low $B_i$ values identify features that capture salient morphological contrasts between domains, high $B_i$ values identify the least discriminative descriptors. These high-Borda-score features are prioritized by the framework as they represent domain-neutral structural properties, serving as the primary candidates for stable, invariant structural anchors.

\paragraph{Step 2: Pairwise manifold adaptation and IIT Score.}
To optimize knowledge propagation between specific source--target combinations, the framework calculates a directed \textit{Importance Inversion Transfer Score} ($\text{IIT}_{\text{score}}^{S,T}$). For brevity, this metric is henceforth referred to as the \emph{IIT score} throughout the manuscript. This metric identifies the optimal structural bridgeheads by integrating global discriminative neutrality with pair-specific rank-order consistency and metric similarity:

\begin{equation}
    \text{IIT}_{\text{score}}^{S,T} (f_i) = B_i \times \frac{\rho_i^{S,T}}{1 + \Delta_i^{S,T}},
\end{equation}

where $f_i$ denotes a specific topological descriptor (feature) from the extracted set, $\rho_i^{S,T}$ represents the Spearman rank correlation coefficient, and $\Delta_i^{S,T} = |\bar{X}_i^S - \bar{X}_i^T|$ denotes the absolute difference between the intra-domain feature means within the standardized manifold. 

The choice of Spearman’s $\rho$ over Pearson’s linear correlation prioritizes \textit{rank-order consistency}, accounting for the non-linear scaling of topological descriptors across disparate generative dynamics. As a non-parametric measure, $\rho$ ensures that the selection remains robust to the skewed distributions and ``physiological'' outliers inherent in real-world relational data. Simultaneously, the denominator functions as an \textit{Inverse Mean Difference} (IMD) penalty. By prioritizing descriptors that exhibit maximal proximity between distributions, the IMD term mitigates \textit{covariate shift}, ensuring that selected features require minimal geometric transformation during manifold synchronization.

The framework identifies the eight descriptors ($f_i$) with the highest IIT score values as the primary structural anchors for a specific directed pair. To ensure system-wide robustness, a hierarchical fallback mechanism is implemented: in scenarios where pairwise alignment is numerically ill-conditioned or yields insufficient signal, the framework defaults to the \textit{general structural anchors}. These global anchors represent the features with the highest \emph{Global Consensus IIT score} ($\text{IIT}_{\text{score, G}}$)---defined as the average IIT score across all investigated domain pairs (see Fig.~\ref{fig:global_ranking}). This hierarchical strategy ensures that the transfer process remains anchored to a robust structural backbone, effectively insulating knowledge propagation from domain-specific idiosyncrasies and reducing the risk of negative transfer. 

\paragraph{Step 3: Manifold synchronization via PCA-SVD alignment.}
Finally, manifold synchronization utilizes Principal Component Analysis (PCA), implemented through Singular Value Decomposition (SVD), to align the feature spaces of the source and target domains. Such subspace alignment represents an established methodology in domain adaptation literature, frequently employed to mitigate distributional shift by projecting disparate datasets into a shared latent space \cite{fernando2013unsupervised, pan2010domain}. Conventional approaches, such as Correlation Alignment (CORAL) or Transfer Component Analysis (TCA), typically apply these transformations to high-dimensional latent embeddings or the entire available feature set to minimize statistical divergence \cite{sun2016deep}. 

The X-CDTL framework introduces a critical distinction: rather than performing alignment on unrefined or opaque representations, the framework restricts the synchronization exclusively to the eight structural anchors identified through the IIT logic. This procedure projects the manifolds into a common orthogonal subspace spanned by the leading singular vectors of their joint covariance matrix. By grounding the synchronization in domain-invariant topological commonalities rather than idiosyncratic noise, the framework ensures a geometrically synchronized mapping that remains theoretically interpretable and robust across scientific boundaries. This selective alignment effectively prevents the ``feature pollution'' often encountered in standard manifold synchronization. As demonstrated in our comparative analysis (see Supplementary Sec.~\ref{subsec:parsimony_evaluation}), restricting the alignment to these structural anchors provides a more stable discriminative grammar for anomaly detection than standard methods that utilize unrefined feature sets, effectively neutralizing the distributional noise that typically leads to negative transfer.

The convergence of these three hierarchical steps ensures that the selected structural anchors remain discriminative, invariant, and geometrically synchronized, establishing a robust foundation for knowledge transfer across heterogeneous scientific domains.

\subsection{Definition of Anomalies and Target Labels}
\label{subsec:anomaly_detection}

A composite extremeness index, $S_i$, defines structural anomalies by quantifying the multivariate structural deviation of a graph from the domain typical topological profile:

\begin{equation}
    S_i = \sum_{j=1}^{k} |z_{ij}|,
\end{equation}

where $z_{ij}$ denotes the $z$-score of feature $f_j$ for graph $i$. The 90th percentile of this index within each domain establishes the decision threshold for anomaly labeling, designating the most extreme 10\% of the population as anomalous. This criterion ensures a domain-agnostic, topology-driven definition of outliers that remains consistent across disparate generative dynamics.

The unsupervised anomaly detection pipeline utilizes the Isolation Forest (iForest) algorithm~\cite{liu2008isolation}. 
Unlike traditional proximity or density-based methods that characterize ``normal'' instances to identify deviations \cite{breunig2000lof, chandola2009anomaly, ramaswamy2000efficient}, iForest explicitly isolates anomalies by exploiting two fundamental properties: they are few in number and possess attribute values significantly different from those of normal instances. 
The algorithm constructs an ensemble of isolation trees ($iTrees$) through recursive random partitioning. In this hierarchical structure, anomalies typically reside closer to the root of the tree, as their distinct topological signatures require fewer random splits to achieve isolation.

Several factors justify the selection of iForest for cross-domain knowledge transfer. Firstly, iForest operates without assuming a specific underlying distribution, a critical requirement given the diverse and often skewed distributions of network descriptors, such as power-law degree patterns. Secondly, the algorithm maintains predictive performance in the multi-dimensional feature space of X-CDTL without the ``curse of dimensionality'' that frequently degrades distance-based metrics. Finally, the ensemble architecture effectively captures complex, non-linear dependencies between structural invariants across disparate scientific domains.

Setting the \textit{contamination} parameter to $0.1$ synchronizes the model decision threshold with the ground-truth generation logic. This configuration ensures that the anomaly score threshold aligns with the 90th percentile of topological deviation, providing a rigorous benchmark for evaluating the stability and recovery capacity of the transfer learning framework under stochastic corruption. 

\subsection{Cross-Domain Transfer Learning Scenarios}

The evaluation of cross-domain knowledge propagation relies on two primary experimental conditions. These scenarios assess how structural priors from a data-rich source domain facilitate anomaly detection in a data-scarce target environment, even when domains exhibit fundamentally different generative dynamics (e.g., transferring principles from dense \textit{Social ego-networks} to sparse \textit{Molecular graphs}).

To ensure a rigorous comparison and isolate the functional contribution of source-domain knowledge from the geometric effects of manifold synchronization, both conditions operate within the same aligned subspace. Specifically, both source and target descriptors undergo the identical PCA-SVD synchronization (Sec.~\ref{subsubsec:steps}, Step 3).

\begin{itemize}
    \item[] \textbf{No Transfer (Aligned Baseline):} This scenario establishes a control by training Isolation Forests exclusively on a limited and corrupted subset of target data. Although the features are projected into the synchronized manifold, the model lacks external structural knowledge. In this regime, the detector performance is strictly bounded by the quality and volume of available target-specific information, making the decision boundary highly susceptible to stochastic threshold collapse under noise.
    
    \item[] \textbf{X-CDTL Transfer (Hybrid Knowledge Integration):} This condition represents the proposed framework, where the model integrates the robust structural backbone of an uncorrupted source domain with the available target samples. By training the Isolation Forest on the union of the synchronized source dataset and the limited target samples, the framework leverages the source connectivity patterns as a \emph{structural skeleton} to regularize the target manifold. 
\end{itemize}

All detectors utilize identical hyperparameters and feature spaces to ensure that performance gains arise strictly from knowledge transfer rather than model complexity. This strategy effectively anchors the learning process to the stable structural invariants identified via the $\text{IIT}_{\text{score}}$, preventing the model from overfitting to local stochastic noise and providing a robust discriminative logic that remains functional even when the target signal is most severely compromised. 

\subsection{Performance Evaluation and Robustness Analysis}

\subsubsection{Model Assessment and Performance Metrics}
\label{subsubsec:TL_metrics_TGI}
Model performance assessment relies on three complementary metrics to capture distinct aspects of classification efficacy: Area Under the Receiver Operating Characteristic curve (ROC-AUC), Average Precision (AP), and the F1-score. ROC-AUC serves as the primary reference due to its threshold-independence and stability under class imbalance, while AP and F1-score provide insights into early precision and decision-boundary sensitivity, respectively. 

The \textit{Transfer Gain Index} (TGI) quantifies the realized functional benefit of knowledge propagation by measuring the relative improvement of a transfer-enabled model ($M_{\text{T}}$) over a target-only baseline ($M_{\text{NT}}$). For a generic performance metric $M \in \{\text{ROC-AUC, AP, F1}\}$, the index is defined as:
\begin{equation}
    \text{TGI}_M(s,t) = \frac{M_{\text{T}} - M_{\text{NT}}}{M_{\text{NT}} + \epsilon},
\end{equation}
where $\epsilon = 10^{-6}$ ensures numerical stability. Together with the $\text{IIT}_{\text{score}}$ (Sec.~\ref{subsubsec:steps}), these metrics support a mechanistic interpretation of transfer learning, distinguishing inherent structural isomorphism from empirical performance gains.

\subsubsection{Multi-Factor Stress-Test Framework}

Knowledge propagation efficacy remains strictly contingent upon target data volume and the scaling properties of the source-domain representation~\cite{caccia2023towards}. Under regimes of extreme data scarcity, maintaining model robustness requires an optimal balance between pre-training depth and fine-tuning adaptation. To evaluate this balance, a systematic perturbation framework assesses transfer resilience under severe data degradation, simulating the suboptimal signal-to-noise ratios and label sparsity inherent in real-world scientific datasets.

The experimental architecture utilizes a multi-factor grid to delineate the framework operational boundaries and resistance to distributional shifts. This design evaluates domain-aligned transfer across 12 directed domain pairs, resulting in an analytical matrix of 90 stochastic realizations per pair. The stress test follows three orthogonal dimensions of variability:

\begin{enumerate}
    \item[] \textbf{Stochastic Realizations ($N=10$):} Ten independent random seeds $\{1, 24, 42, 50, 123, 501, 700, 800, 920, 999\}$ mitigate sampling bias and ensure high statistical power. For each seed, a resampling procedure extracts a subset of 500 graphs from master pools of 1,000 instances per domain, ensuring that results represent independent realizations of the underlying graph populations.
    \item[] \textbf{Data Scarcity Regimes ($\alpha \in \{0.1, 0.5, 0.9\}$):} This parameter defines the sampling fraction, or the proportion of target instances available for model adaptation. Lower $\alpha$ values test the capacity of source-domain structural anchors to regularize the decision manifold when local information is insufficient for independent learning.
    \item[] \textbf{Feature Corruption Levels ($\eta \in \{0.1, 0.5, 0.9\}$):} Resiliency is characterized via the injection of additive Gaussian noise, $\mathcal{N}(0, \eta^2)$. Scaling the noise magnitude according to the feature-wise standard deviation ensures comparability across heterogeneous topological descriptors. The protocol further incorporates a fixed missing-value fraction ($0.05$) to account for non-computable metrics, utilizing median imputation to preserve numerical stability.
\end{enumerate}

To ensure ecological validity, the pipeline restricts these perturbations exclusively to the training phase while evaluating models on an uncorrupted held-out set. This simulates a realistic deployment scenario where a model trained on low-quality data must generalize to high-fidelity observations. For every coordinate in the grid, the pipeline compares the baseline ``No Transfer (Aligned)'' scenario against the ``X-CDTL Transfer'' approach. 

Computational integrity is maintained through robust safeguards during the manifold alignment stage. If ill-conditioned covariance matrices prevent SVD convergence on the GPU, an automatic CPU-based fallback mechanism activates. This comprehensive stress-test framework provides the empirical basis for quantifying the \textit{rescue effect} provided by structural anchors in highly corrupted environments. 

\subsubsection{Sample Complexity and Dimensionality Control}
\label{subsubsec:dimensionality_control}

The ratio between observations ($N$) and the dimensionality of the feature space ($p$) governs the statistical reliability of the framework. While the initial domain classification task operates with a high sample-to-feature ratio ($N/p \approx 166.7$; $N=2,000, p=12$), transfer experiments adjust $N$ according to the target data fraction $\alpha$.

The selection of eight structural anchors—representing 66\% of the available topological feature space—identifies a \textit{topological sweet spot} that optimizes statistical density. Reducing the feature space from $p=12$ to $p=8$ increases the minimum $N/p$ ratio from $50.0$ to $75.0$, a 50\% improvement in sample density per feature dimension. This dimensionality control preserves the discriminative power of the Isolation Forest engine in low-data regimes. Across the entire experimental grid, the framework operates within a high-confidence statistical zone ($N/p \gg 10$), ensuring that performance gains reflect shared topological principles rather than numerical artifacts or overfitting to idiosyncratic noise.

\subsubsection{Statistical Validation and Meta-Analysis}
A meta-analysis of the results aggregated across the robustness grid determines the statistical significance of performance gains. Non-parametric inference accounts for potential non-normality in performance distributions under stress. Specifically, the Kruskal-Wallis $H$-test evaluates global differences across scenarios, while Dunn’s post-hoc test with Bonferroni correction provides the multi-comparison ranking. 

Aggregation of data from all independent realizations into a high-density dataset facilitates a granular characterization of structural identities. This multi-seed strategy minimizes the impact of sampling noise, ensuring that the comparative analysis of structural motifs remains invariant to the specific stochastic selection of nodes and edges within individual runs. Statistical significance for observed trends in performance shifts is further determined via paired two-tailed $t$-tests.

\subsubsection{Computational Infrastructure and GPU Implementation}

The computational framework integrates a specialized Python ecosystem, utilizing \texttt{networkx} for graph-theoretical operations, \texttt{scikit-learn} for machine learning baselines, and \texttt{PyTorch Geometric} for structured relational data processing. To ensure high throughput during multi-scale feature extraction and manifold synchronization, the pipeline implements a dynamic GPU-selection logic. This system queries hardware availability via \texttt{nvidia-smi} and automatically assigns computations to the most efficient available resource, specifically leveraging high-performance architectures such as the NVIDIA RTX 4090 (24\,GB VRAM) or the NVIDIA A100-PCIE (40\,GB VRAM).

Computational determinism and experimental reproducibility are ensured through fixed random seeds across all stochastic modules (\texttt{numpy}, \texttt{torch}, and \texttt{random}). Preprocessing, normalization, and sampling routines are standardized within a version-controlled environment to guarantee consistent data partitioning and parameter persistence across all 90 stochastic realizations.

The calculation of topological indices and cross-domain similarity measures relies on CUDA-optimized tensor operations within the \texttt{PyTorch} framework. This infrastructure facilitates both numerical precision and the feasibility of performing thousands of feature-level comparisons and repeated manifold alignments. By offloading these operations to GPU kernels, empirical estimation latency is reduced by over one order of magnitude compared to CPU-based implementations, enabling the exhaustive evaluation of transferability and stability metrics across the multi-factor experimental grid.

\backmatter

\section{Supplementary Information}

\subsection{Cross-Domain Structural Landscapes and Ensemble Variability}

\begin{figure}[p] 
    \centering
    
    \begin{subfigure}{0.32\textwidth}
        \includegraphics[width=\textwidth, height=0.18\textheight, keepaspectratio]{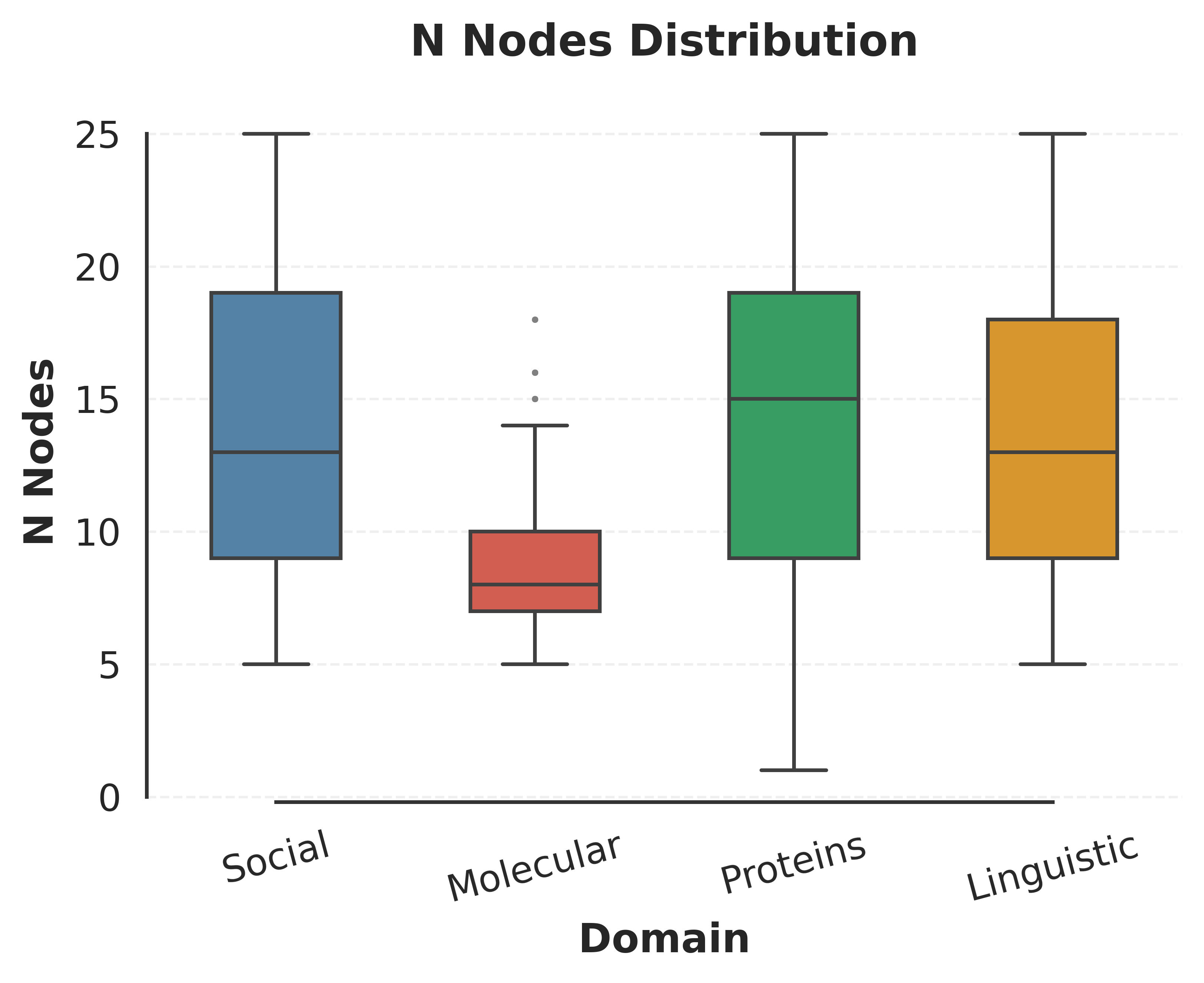}
    \end{subfigure}
    \hfill
    \begin{subfigure}{0.32\textwidth}
        \includegraphics[width=\textwidth, height=0.18\textheight, keepaspectratio]{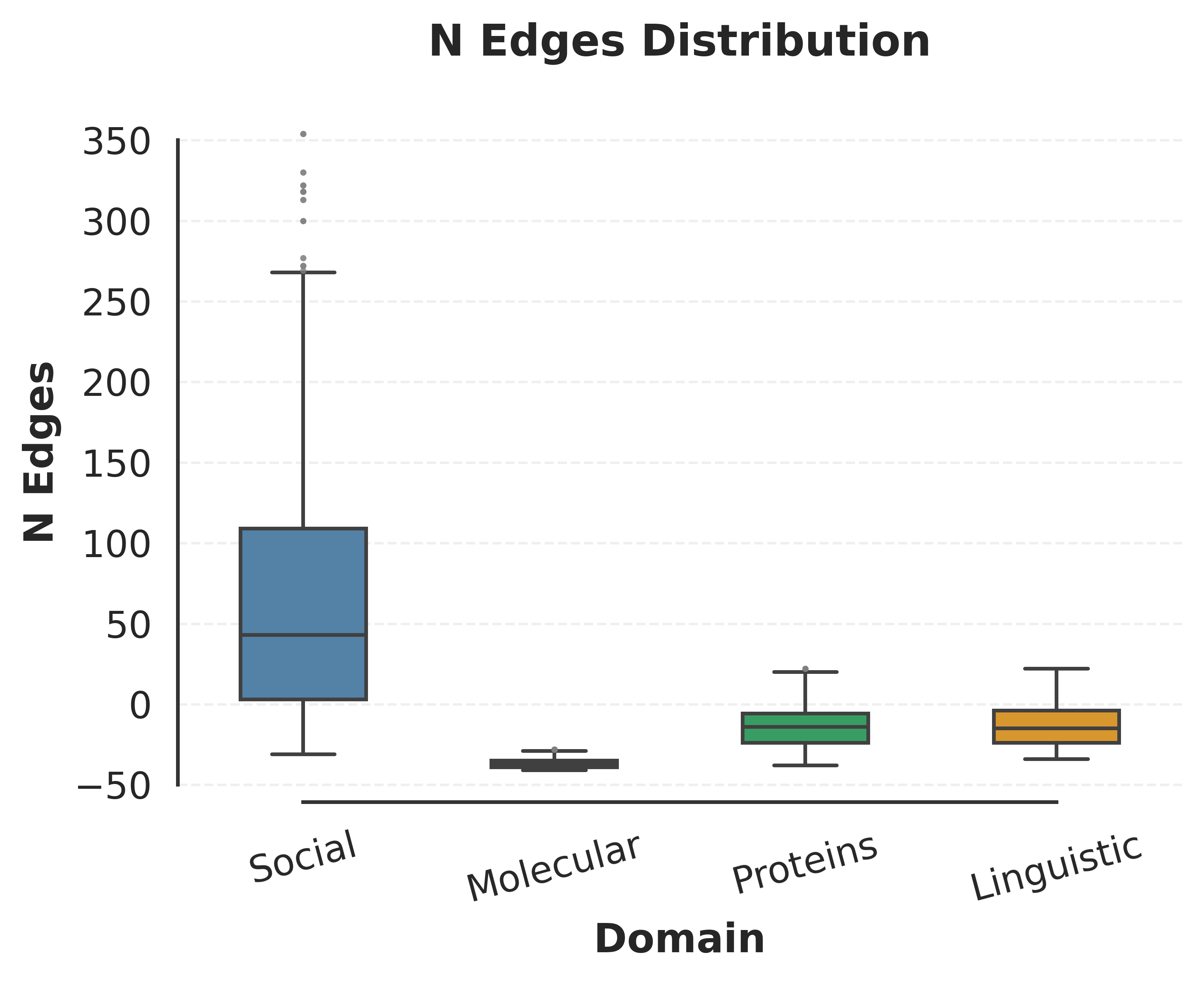}
    \end{subfigure}
    \hfill
    \begin{subfigure}{0.32\textwidth}
        \includegraphics[width=\textwidth, height=0.18\textheight, keepaspectratio]{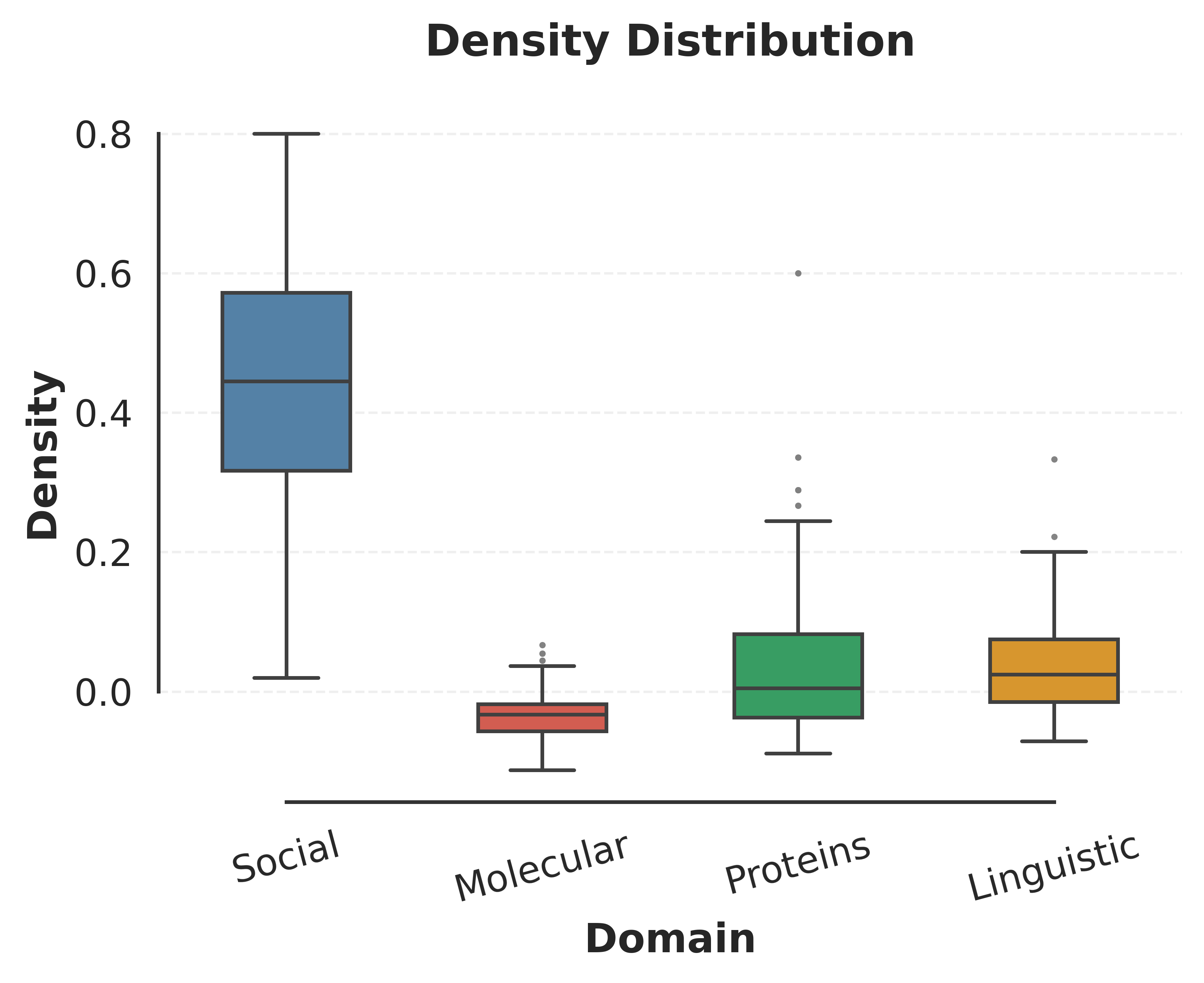}
    \end{subfigure}

    \vskip 5pt 
    
    \begin{subfigure}{0.32\textwidth}
        \includegraphics[width=\textwidth, height=0.18\textheight, keepaspectratio]{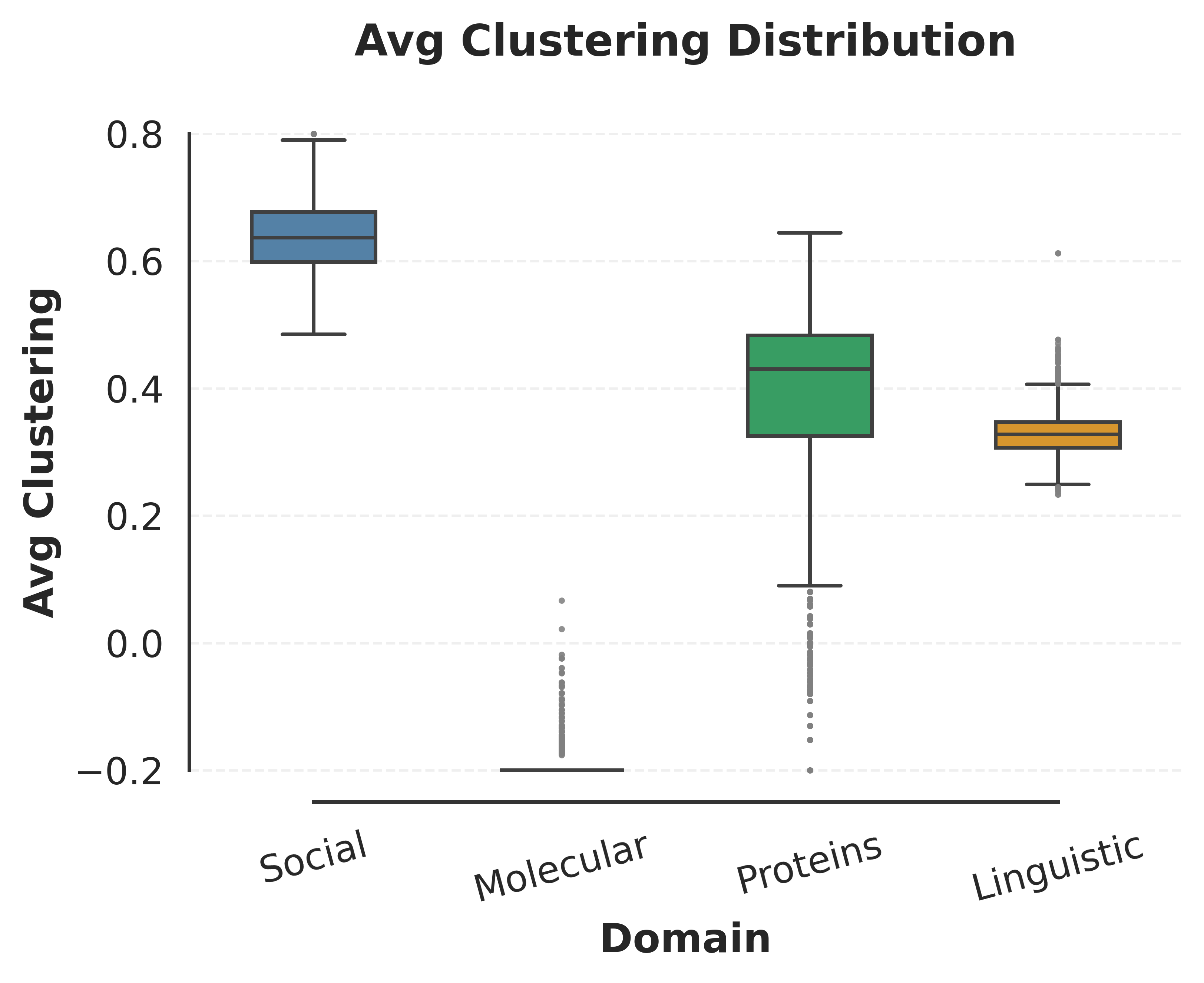}
    \end{subfigure}
    \hfill
    \begin{subfigure}{0.32\textwidth}
        \includegraphics[width=\textwidth, height=0.18\textheight, keepaspectratio]{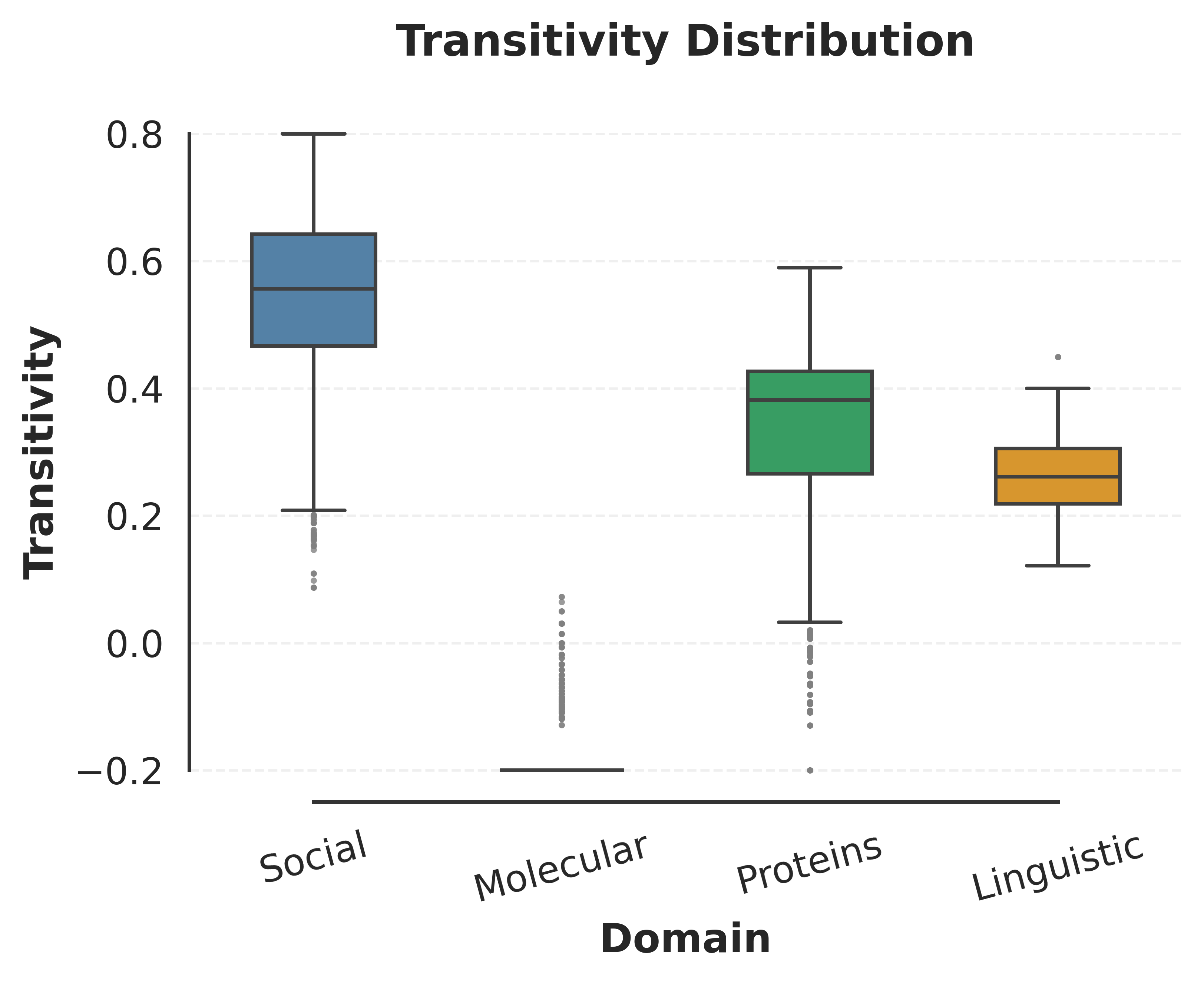}
    \end{subfigure}
    \hfill
    \begin{subfigure}{0.32\textwidth}
        \includegraphics[width=\textwidth, height=0.18\textheight, keepaspectratio]{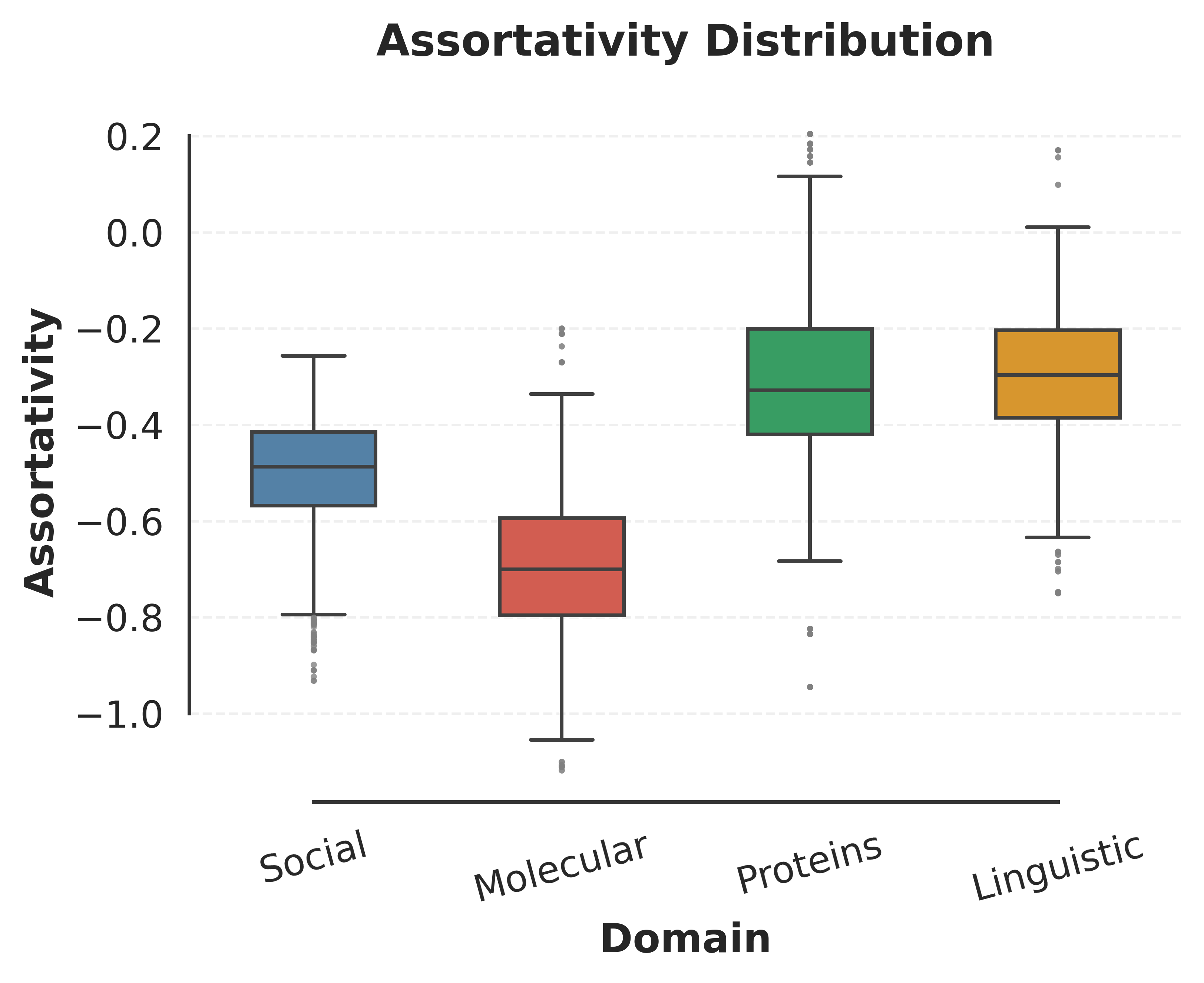}
    \end{subfigure}

    \vskip 5pt

    \begin{subfigure}{0.32\textwidth}
        \includegraphics[width=\textwidth, height=0.18\textheight, keepaspectratio]{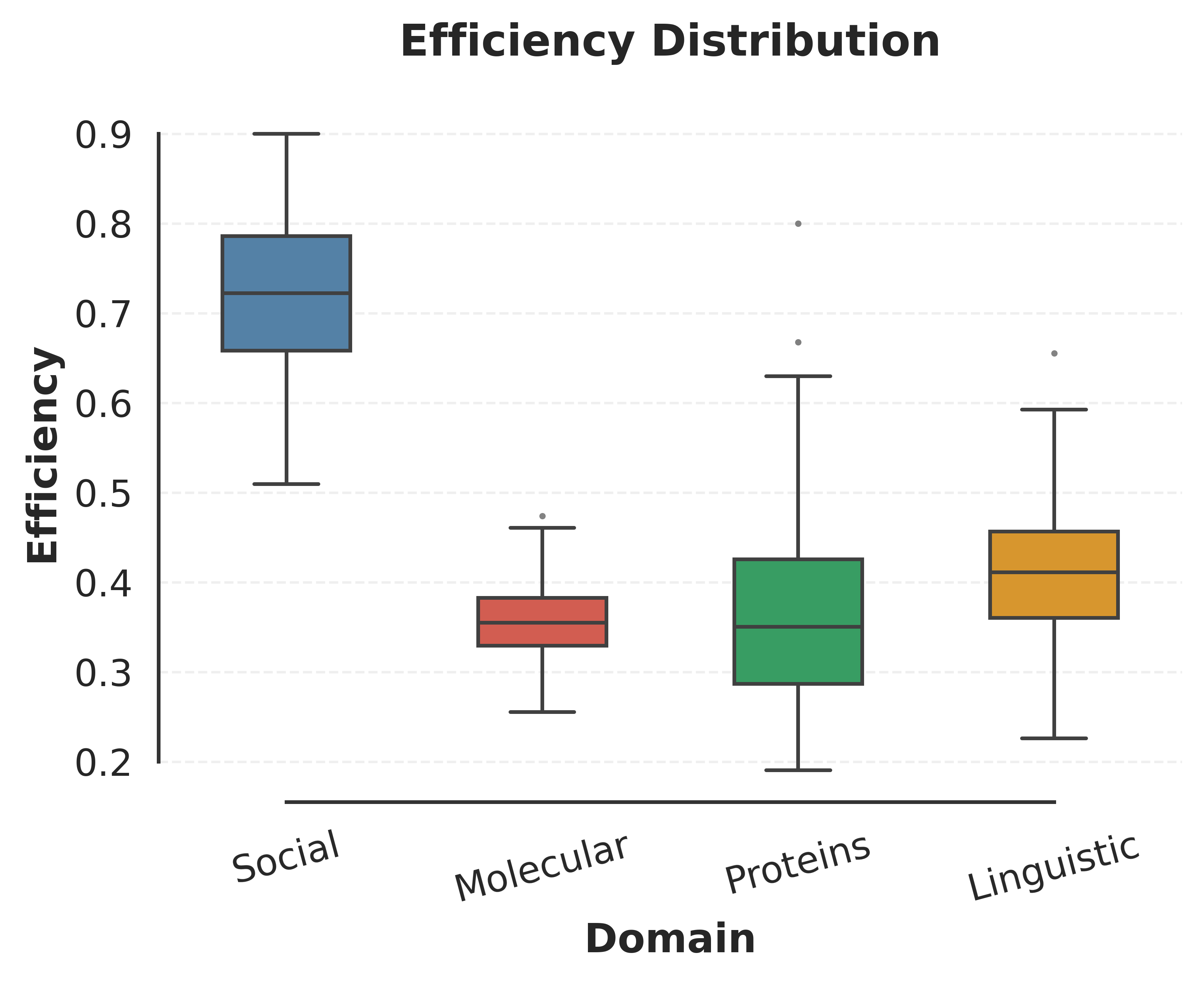}
    \end{subfigure}
    \hfill
    \begin{subfigure}{0.32\textwidth}
        \includegraphics[width=\textwidth, height=0.18\textheight, keepaspectratio]{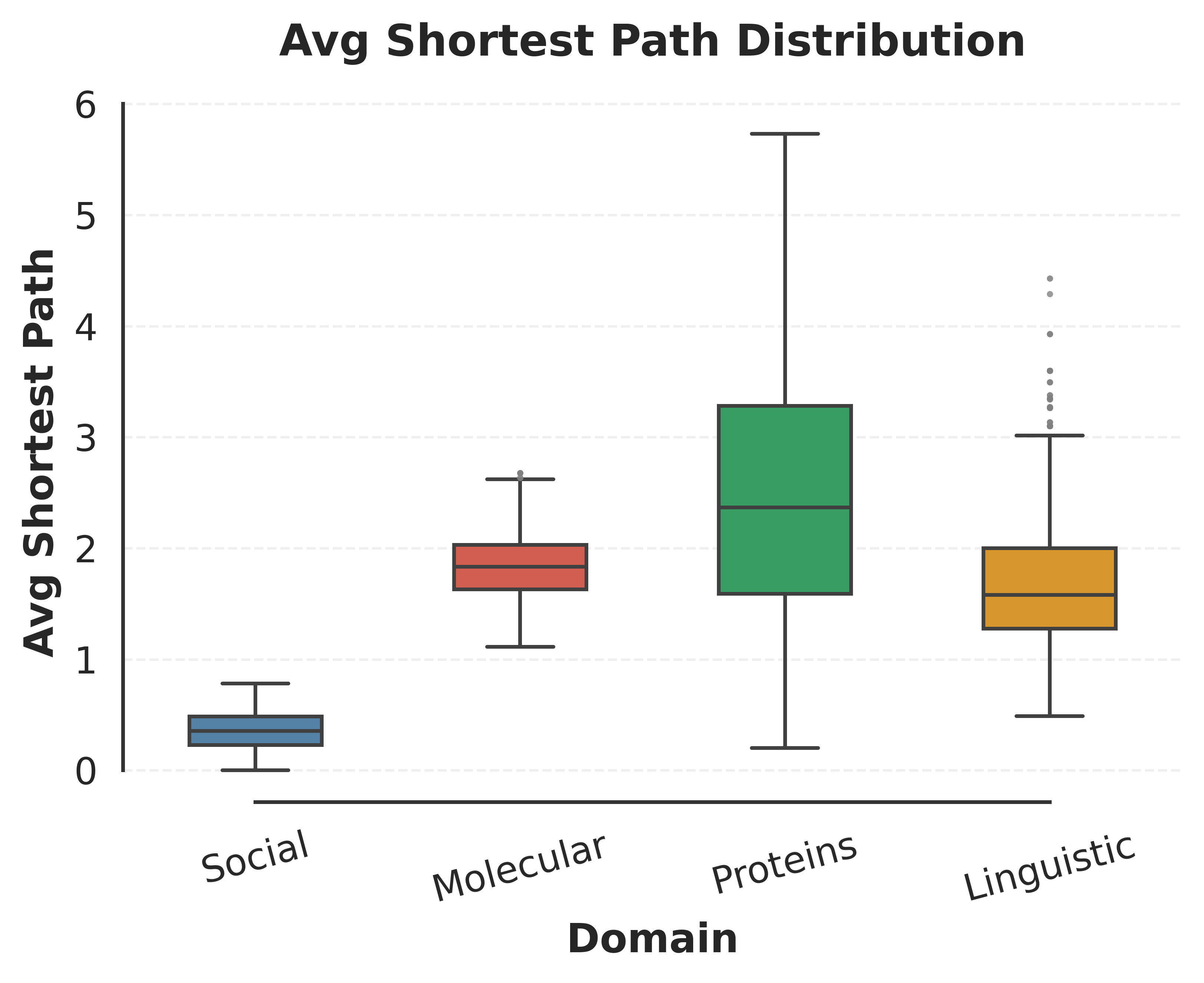}
    \end{subfigure}
    \hfill
    \begin{subfigure}{0.32\textwidth}
        \includegraphics[width=\textwidth, height=0.18\textheight, keepaspectratio]{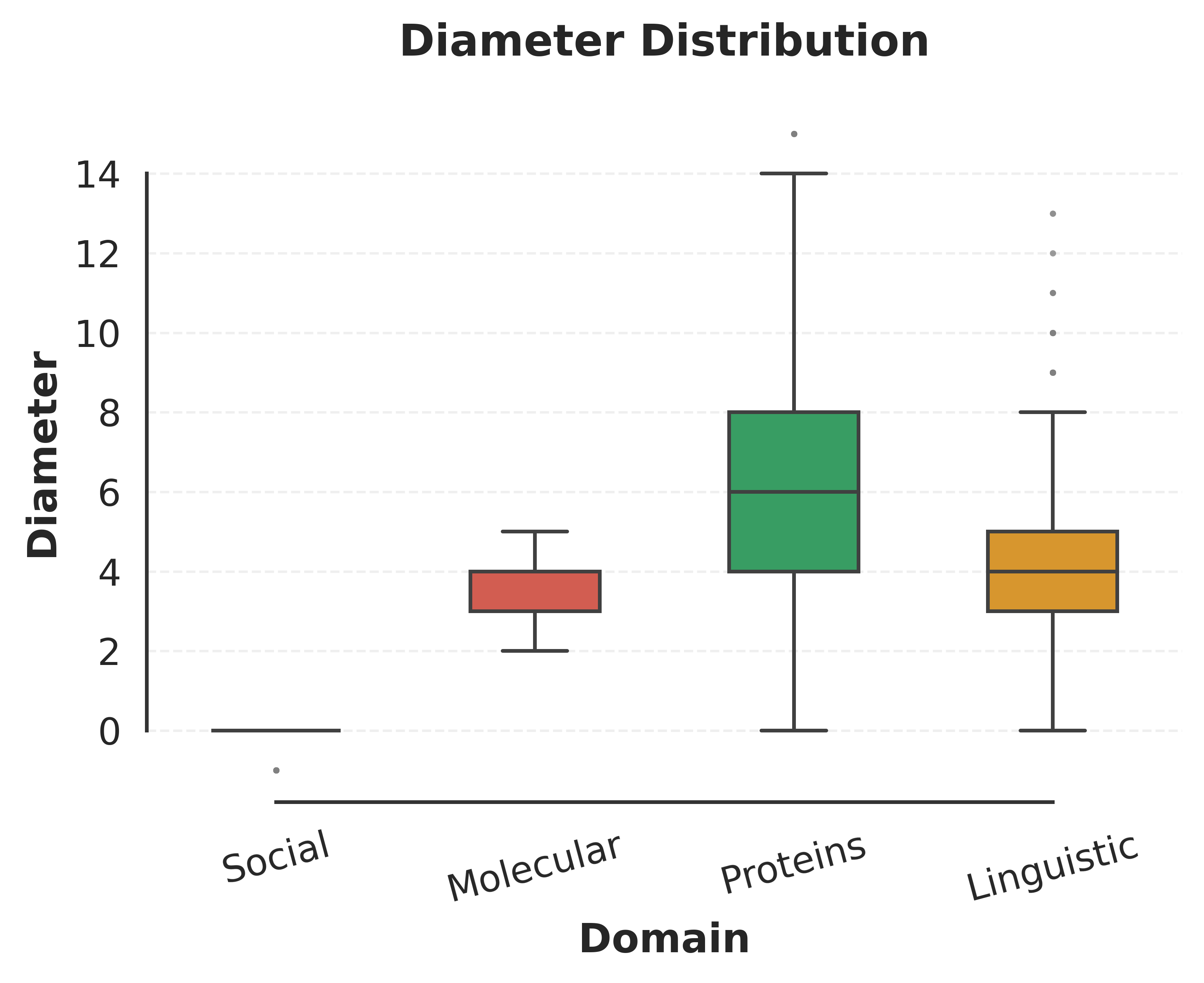}
    \end{subfigure}

    \vskip 5pt

    \begin{subfigure}{0.32\textwidth}
        \includegraphics[width=\textwidth, height=0.18\textheight, keepaspectratio]{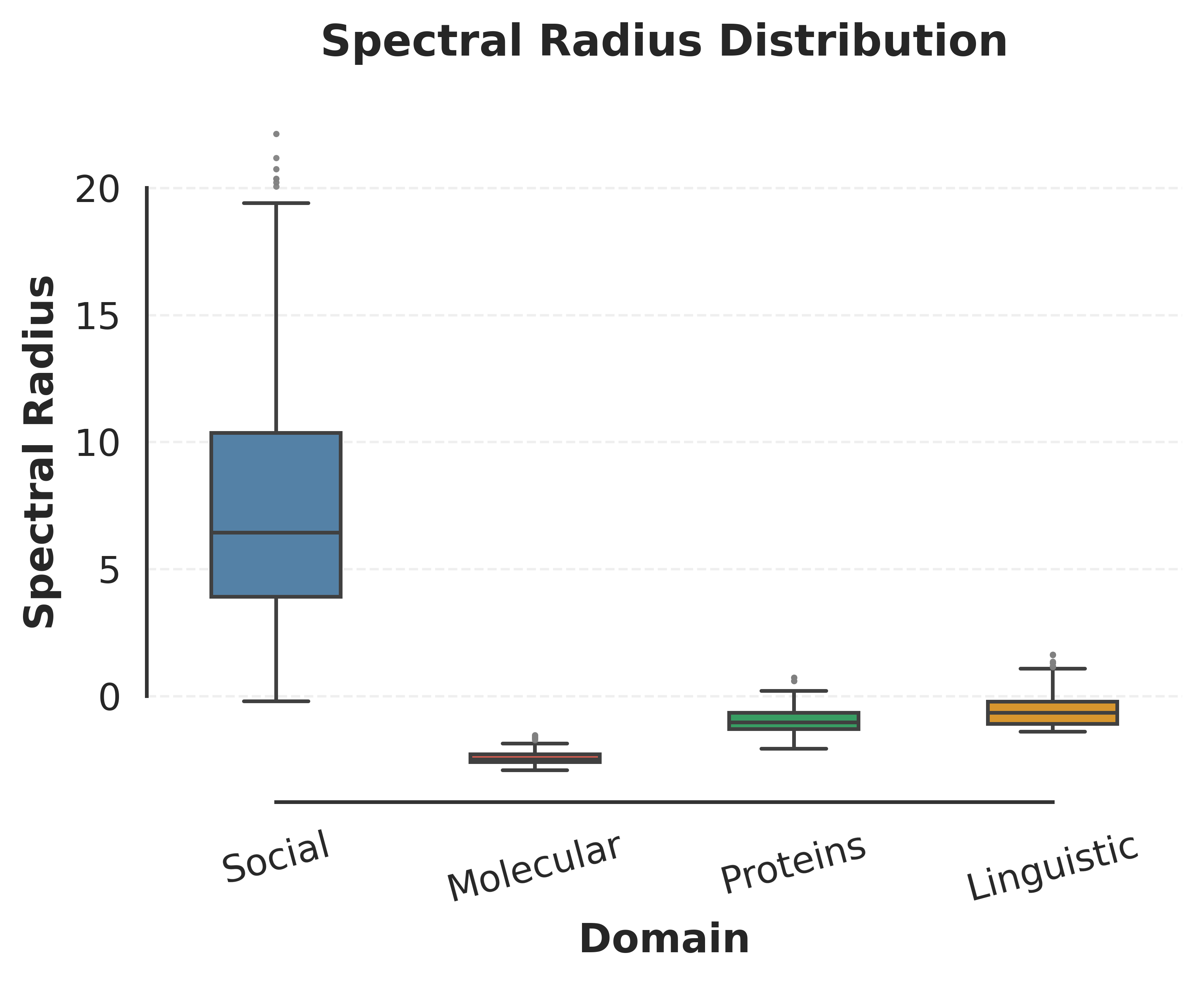}
    \end{subfigure}
    \hfill
    \begin{subfigure}{0.32\textwidth}
        \includegraphics[width=\textwidth, height=0.18\textheight, keepaspectratio]{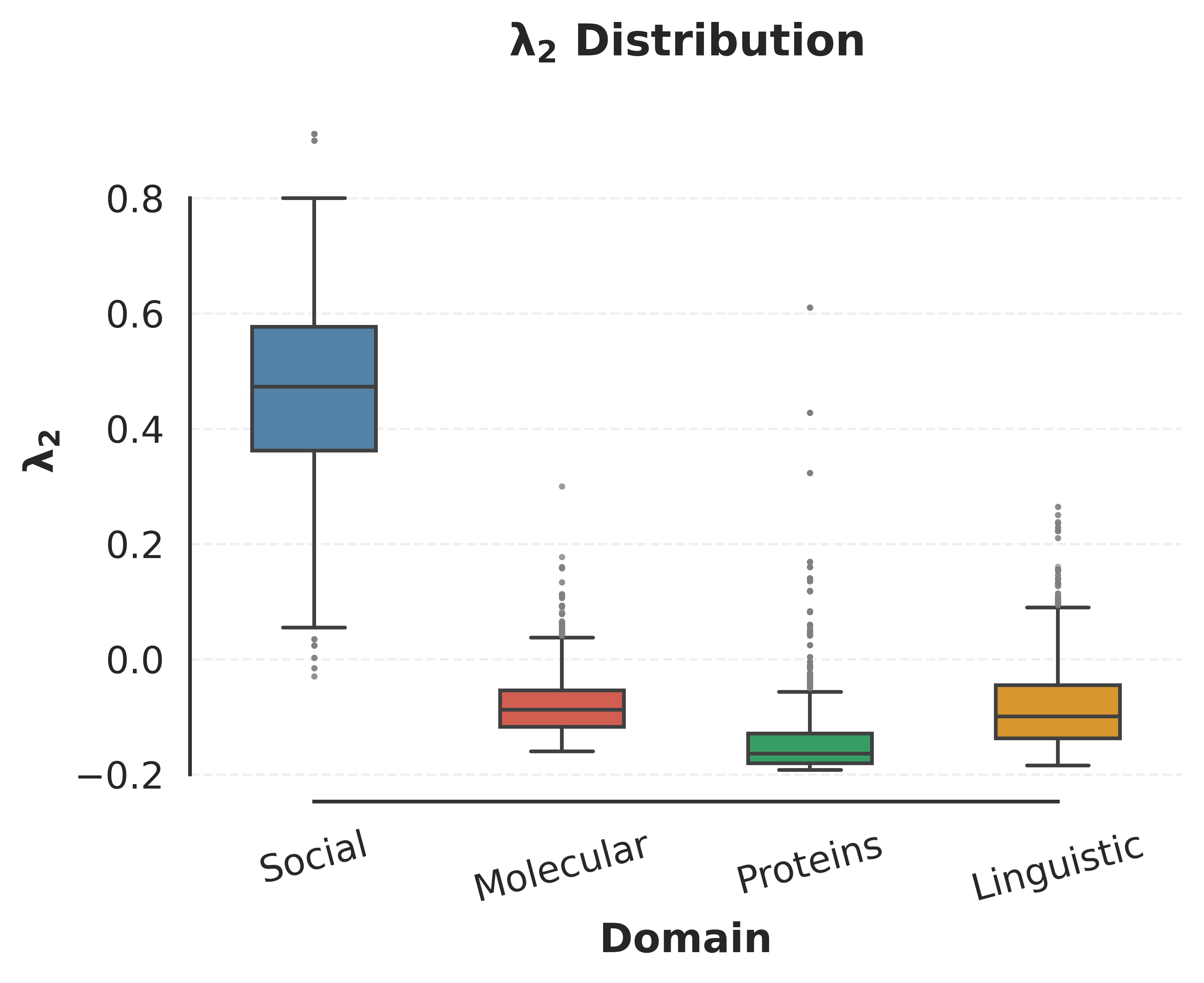}
    \end{subfigure}
    \hfill
    \begin{subfigure}{0.32\textwidth}
        \includegraphics[width=\textwidth, height=0.18\textheight, keepaspectratio]{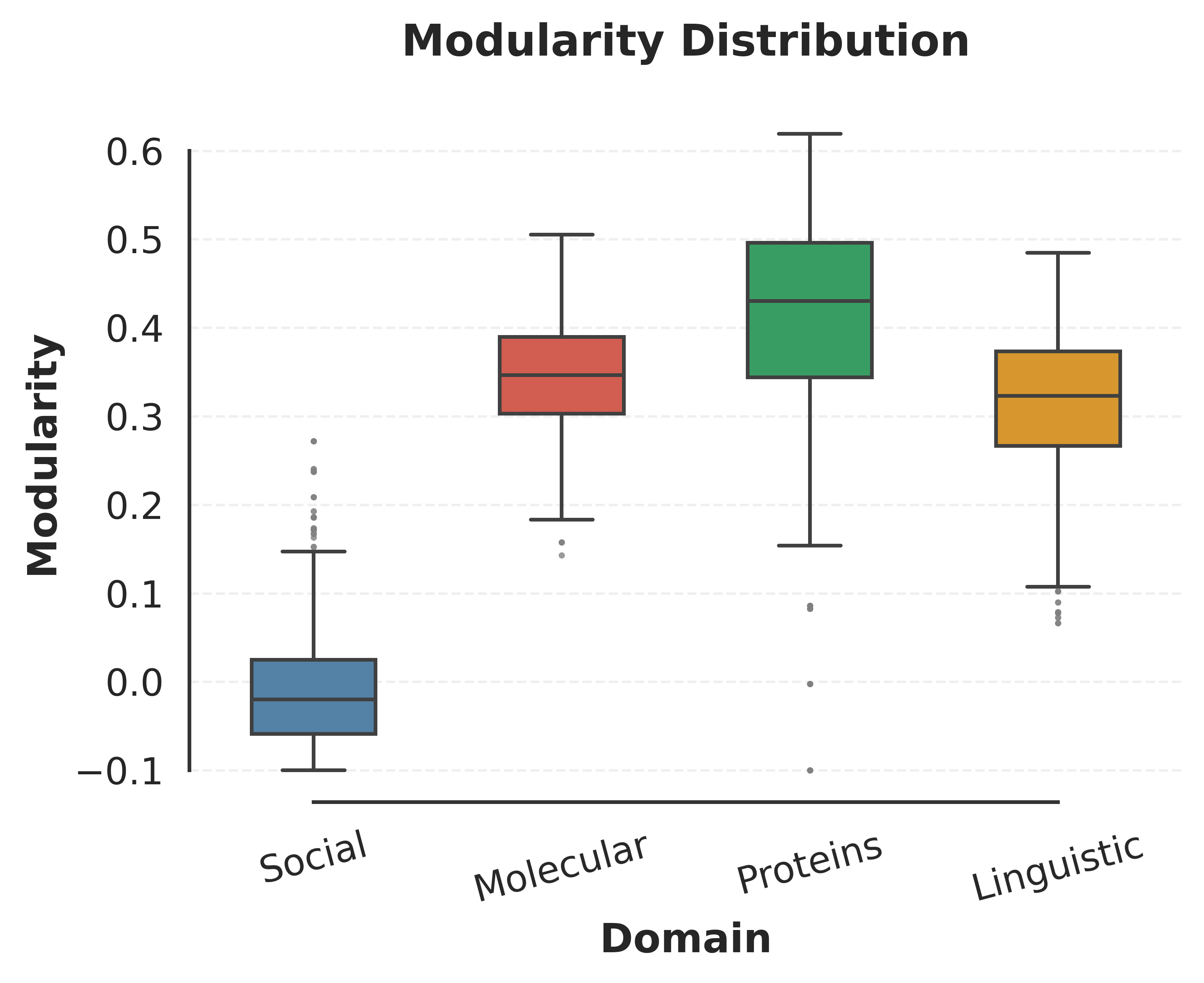}
    \end{subfigure}
    
    \vspace{10pt} 
    
    \caption{\textbf{Multi-scale structural distributions across network domains.} Comparative boxplots for the four graph domains demonstrate the high separability of the structural feature space. The wide range of outliers underscores the inherent complexity of real-world relational data, while the robust non-overlapping distributions for key metrics like Density and Spectral Radius validate the selection of these domains as a demanding benchmark for cross-domain alignment. The pooled dataset of 20,000 graphs ($N=5,000$ per domain) underlies all distributional plots.}
    \label{fig:boxplots_supplementary}
\end{figure}

Ensemble structural statistics, aggregated across 10 independent stochastic realizations, reveal significant intra-domain heterogeneity (Table~\ref{tab:supp_descriptive_stats}). Intrinsic relational variability within real-world data manifests through ``physiological'' outliers across all topological dimensions (Fig.~\ref{fig:boxplots_supplementary}). These fluctuations reflect the complexity of the generative processes involved: social ego-networks, for instance, exhibit long-tailed distributions in edge counts ($113.90 \pm 73.28$), with specific instances reaching up to 404 links. Conversely, the molecular domain contains rare cyclic structures within an otherwise nearly acyclic space, while protein networks display extreme variations in modularity and diameter ($8.12 \pm 3.24$).

\begin{table}[h!]
\centering
\caption{\textbf{Ensemble structural statistics across network domains.} 
Values denote the ensemble mean $\pm$ standard deviation ($\mu \pm \sigma$) aggregated over 10 independent seeds ($N=5,000$ graphs per domain). Bold values identify the primary topological discriminants: the quasi-clique redundancy of social networks, the high modularity of protein interaction networks, and the structural sparsity characteristic of molecular graphs.}
\label{tab:supp_descriptive_stats}
\footnotesize 
\setlength{\tabcolsep}{4pt} 
\renewcommand{\arraystretch}{1.2} 
\begin{tabular}{l cccc}
\toprule
\textbf{Metric} & \textbf{Social} & \textbf{Molecular} & \textbf{Proteins} & \textbf{Linguistic} \\
                & ($\mu \pm \sigma$) & ($\mu \pm \sigma$) & ($\mu \pm \sigma$) & ($\mu \pm \sigma$) \\
\midrule
$n_{\text{nodes}}$       & $18.76 \pm 5.90$   & $13.34 \pm 2.44$  & $19.27 \pm 5.90$  & $18.72 \pm 5.32$  \\
$n_{\text{edges}}$       & $\mathbf{113.90 \pm 73.28}$ & $13.14 \pm 2.44$  & $36.07 \pm 12.02$ & $36.61 \pm 12.69$ \\
Density                  & $\mathbf{0.65 \pm 0.17}$ & $\mathbf{0.17 \pm 0.03}$ & $0.23 \pm 0.09$ & $0.23 \pm 0.06$ \\
Avg. Clustering          & $\mathbf{0.84 \pm 0.06}$ & $\mathbf{0.01 \pm 0.03}$ & $0.58 \pm 0.17$ & $0.53 \pm 0.04$ \\
Transitivity             & $0.74 \pm 0.14$    & $0.03 \pm 0.05$   & $0.53 \pm 0.15$   & $0.46 \pm 0.05$   \\
Assortativity            & $-0.30 \pm 0.12$   & $-0.49 \pm 0.15$  & $-0.10 \pm 0.18$  & $-0.11 \pm 0.13$  \\
Efficiency               & $0.82 \pm 0.09$    & $0.46 \pm 0.04$   & $0.46 \pm 0.10$   & $0.51 \pm 0.07$   \\
Avg. Short. Path         & $1.36 \pm 0.17$    & $2.84 \pm 0.28$   & $3.51 \pm 1.14$   & $2.69 \pm 0.59$   \\
Diameter                 & $1.99 \pm 0.10$    & $5.28 \pm 0.74$   & $\mathbf{8.12 \pm 3.24}$ & $6.14 \pm 1.91$   \\
Spectral Radius          & $\mathbf{12.28 \pm 4.32}$ & $2.56 \pm 0.21$ & $4.04 \pm 0.45$ & $4.41 \pm 0.57$ \\
$\lambda_2$ (Alg. Conn.) & $0.66 \pm 0.16$    & $0.12 \pm 0.06$   & $0.06 \pm 0.07$   & $0.12 \pm 0.07$   \\
Modularity               & $0.09 \pm 0.07$    & $0.45 \pm 0.06$   & $\mathbf{0.52 \pm 0.11}$ & $0.42 \pm 0.07$   \\
\bottomrule
\end{tabular}
\end{table}

The X-CDTL framework maintains robustness against this heterogeneity through median-based imputation and non-parametric statistical inference. High consistency in structural signatures across independent seeds confirms that global domain identities remain stable despite local stochasticity. This stability ensures that the framework reliance on shared topological laws persists even when individual samples deviate significantly from the ensemble mean.

\subsection{Classification Performance and Topological Discriminants}

\begin{figure}[ht]
    \centering
    \includegraphics[width=0.6\textwidth]{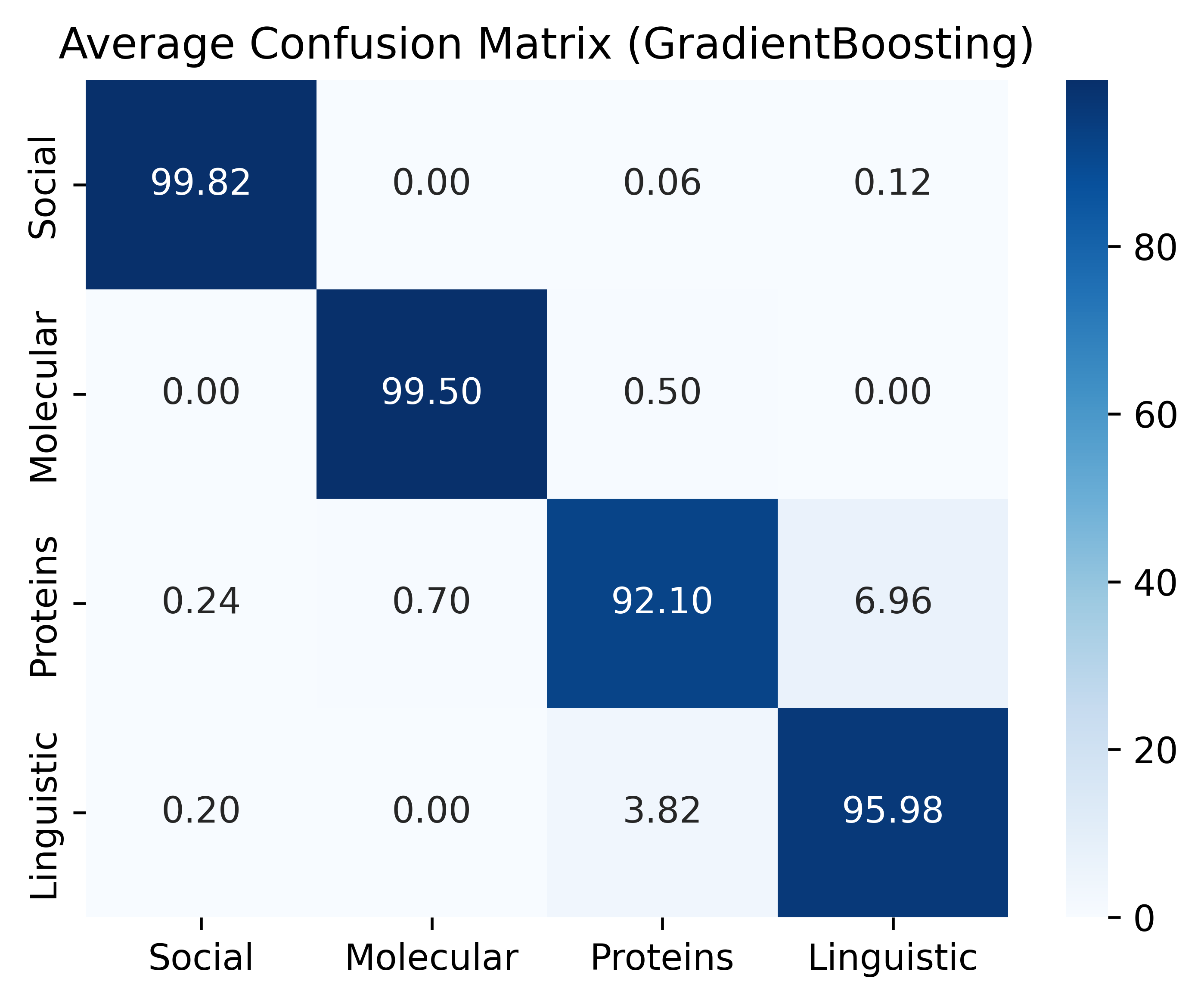} 
    \caption{\textbf{Domain classification performance.} Normalized confusion matrix averaged over 10 independent random seeds for the Gradient Boosting classifier. The extreme diagonal dominance ($>99\%$ for Social and Molecular) confirms the robustness of topological fingerprints, while the modest overlap between Proteins and Linguistic networks ($6.96\%$) highlights shared hierarchical organizational principles in informational systems.}
    \label{fig:fig_conf_mat_supp}
\end{figure}

Supervised classification diagnostics confirm that standardized topological metrics function as high-fidelity fingerprints of network identity, establishing distinct structural regimes across the investigated domains. The ensemble confusion matrix for the Gradient Boosting model (Fig.~\ref{fig:fig_conf_mat_supp}) exhibits acute diagonal dominance, validating the existence of highly separable morphological manifolds. 

In particular, the \textit{Social} and \textit{Molecular} domains achieve near-perfect classification accuracies of $99.82\%$ and $99.50\%$, respectively. This suggests that their generative dynamics (dense triadic redundancy for social networks and strict chemical valence for molecules) produce unambiguous topological signatures. Conversely, Fig.~\ref{fig:fig_conf_mat_supp} reveals that the primary source of misclassification occurs between \textit{Protein} and \textit{Linguistic} networks. Specifically, $6.96\%$ of protein networks are classified as linguistic, while $3.82\%$ of linguistic networks are identified as protein-based. This directed overlap identifies a shared hierarchical organization common to information-bearing networks, where modular sub-structures---such as functional protein domains and syntactic word clusters---facilitate distributed information flow.

Aggregation of feature importance across the three machine learning architectures via a Borda count strategy (Table~\ref{tab:borda_importance}) identifies the primary topological discriminants driving domain separability. Metrics associated with local cohesion (\textit{average clustering coefficient}), spectral dominance (\textit{spectral radius}), and global integration (\textit{diameter}) consistently occupy the highest importance ranks. While these features are essential for high-fidelity domain identification, their prominence indicates significant morphological volatility across scientific disciplines. 

Specifically, descriptors such as \textit{avg\_clustering} (Borda score: 2.267) and \textit{diameter} (4.333) capture domain-specific generative constraints that are poorly preserved under cross-domain shifts. Consequently, as detailed in the hierarchical selection protocol, these primary discriminants are prioritized for exclusion in the IIT mechanism. By filtering out these volatile markers, the framework instead isolates more stable, domain-invariant structural anchors---such as \textit{n\_nodes} (10.700) and \textit{assortativity} (8.767)---to support robust knowledge propagation across heterogeneous systems.

\begin{table}[ht]
\centering
\caption{\textbf{Consensus feature importance ranking (Borda count).} Topological descriptors are ranked from highest (1) to lowest (12) importance across Random Forest (RF), Gradient Boosting (GB), and Logistic Regression (LR) architectures. Metrics represent ensemble averages over 10 independent random seeds. Lower Borda scores indicate higher discriminative utility and cross-architecture consensus for domain identification. These high-importance features represent the primary discriminants that the IIT mechanism subsequently inverts to isolate stable, domain-neutral structural anchors for cross-disciplinary transfer learning.}
\label{tab:borda_importance}
\footnotesize
\renewcommand{\arraystretch}{1.2}
\begin{tabular}{lcccc}
\toprule
\textbf{Feature} & \textbf{RF Rank} & \textbf{GB Rank} & \textbf{LR Rank} & \textbf{Borda Score} \\ 
\midrule
avg\_clustering     & 1.0  & 2.8  & 3.0  & 2.267  \\
spectral\_radius    & 2.3  & 1.7  & 8.7  & 4.233  \\
diameter            & 4.0  & 2.0  & 7.0  & 4.333  \\
modularity          & 5.6  & 6.3  & 1.8  & 4.567  \\
transitivity        & 2.7  & 4.9  & 10.0 & 5.867  \\
avg\_shortest\_path & 6.1  & 9.8  & 4.9  & 6.933  \\
density             & 11.0 & 9.4  & 1.2  & 7.200  \\
lambda\_2           & 6.3  & 3.9  & 12.0 & 7.400  \\
n\_edges            & 8.5  & 7.9  & 6.0  & 7.467  \\
efficiency          & 9.7  & 11.0 & 4.1  & 8.267  \\
assortativity       & 8.8  & 6.5  & 11.0 & 8.767  \\
n\_nodes            & 12.0 & 11.8 & 8.3  & 10.700 \\
\bottomrule
\end{tabular}
\end{table} 

\subsection{Evaluation of Feature Parsimony and Dimensionality Control}
\label{subsec:parsimony_evaluation}

A global comparison between the parsimonious set of eight structural anchors (\textit{Top Feats}) and the complete topological descriptor set (\textit{All Feats}) quantifies the functional impact of dimensionality control (Table~\ref{tab:supp_regime_comparison}). Analysis identifies a fundamental trade-off between structural resolution and manifold alignment stability. While the high-dimensional \textit{All Feats} configuration leverages greater resolution to achieve superior absolute scores in specific high-noise/high-data instances (Transfer F1 $= 0.606$ vs. $0.532$), the \textit{Top Feats} configuration remains the more robust and reliable strategy across the broader experimental landscape.

\begin{table}[h!]
\centering
\caption{\textbf{Comparative Analysis of Feature Parsimony and Transfer Robustness.} Marginalized averages for ROC-AUC, Average Precision (AP), and F1-score across operational regimes. Bold values highlight the performance of the optimized structural anchors (\textit{Top Feats}) when they either outperform the high-dimensional configuration (\textit{All Feats}) in transfer scenarios or achieve a successful rescue effect over the target-only baseline (T $>$ NT).}
\label{tab:supp_regime_comparison}
\small
\setlength{\tabcolsep}{4pt}
\renewcommand{\arraystretch}{1.2}
\begin{tabular}{ll cc cc cc}
\toprule
\textbf{Regime} & \textbf{Feat. Set} & \multicolumn{2}{c}{\textbf{ROC-AUC}} & \multicolumn{2}{c}{\textbf{AP}} & \multicolumn{2}{c}{\textbf{F1-score}} \\
\cmidrule(lr){3-4} \cmidrule(lr){5-6} \cmidrule(lr){7-8}
& & \textbf{NT} & \textbf{T} & \textbf{NT} & \textbf{T} & \textbf{NT} & \textbf{T} \\
\midrule
\multirow{2}{*}{Full Exp. Grid} & All Feats & 0.987 & 0.977 & 0.918 & 0.886 & 0.665 & 0.556 \\
 & Top Feats & 0.987 & 0.974 & 0.917 & 0.879 & 0.660 & \textbf{0.559} \\
\midrule
\multirow{2}{*}{High-Purity ($\eta=0.1, \alpha=0.1$)} & All Feats & 0.983 & 0.958 & 0.893 & 0.812 & 0.747 & 0.292 \\
 & Top Feats & 0.983 & 0.953 & 0.896 & \textbf{0.819} & 0.760 & \textbf{0.312} \\
\midrule
\multirow{2}{*}{High-Noise ($\eta=0.9, \alpha=0.9$)} & All Feats & 0.991 & 0.988 & 0.940 & 0.930 & 0.360 & 0.606 \\
 & Top Feats & 0.990 & 0.985 & 0.933 & 0.916 & 0.341 & \textbf{0.532} \\
\bottomrule
\end{tabular}
\end{table}

Notably, in transfer scenarios marginalized over the full experimental grid, the optimized structural anchors achieve a higher global F1-score ($0.559$) than the complete feature set ($0.556$). This performance gain confirms that the targeted exclusion of unstable discriminants---governed by the \emph{Global Consensus IIT$_{\text{score, G}}$}---prevents ``feature pollution'' during manifold synchronization. This advantage is further amplified in the high-purity regime ($\eta = 0.1, \alpha = 0.1$), where parsimonious selection limits distributional interference to yield performance gains across the metric spectrum. Specifically, the optimized structural anchors achieve both a higher AP ($0.819$ vs. $0.812$) and F1-score ($0.312$ vs. $0.292$) compared to the high-dimensional configuration.

Most significantly, in high-noise environments ($\eta = 0.9, \alpha = 0.9$), the \textit{Top Feats} configuration facilitates a substantial rescue effect, elevating the F1-score from a baseline of $0.341$ to $0.532$ (Table~\ref{tab:supp_regime_comparison}). These findings demonstrate that while higher feature resolution can be beneficial under specific data-rich conditions, the prioritization of domain-invariant anchors is the primary driver of reliable generalization. By filtering out idiosyncratic volatility, the framework ensures that knowledge propagation remains anchored to a stable structural backbone, effectively mitigating the decision-boundary collapse typically observed when local target signals are most severely compromised. 

\subsection{Regime-Specific Stability and Affinity Dynamics}
\label{subsec:affinity_gain_dynamics}

\begin{figure}[H]
    \centering
    \includegraphics[width=0.65\textwidth]{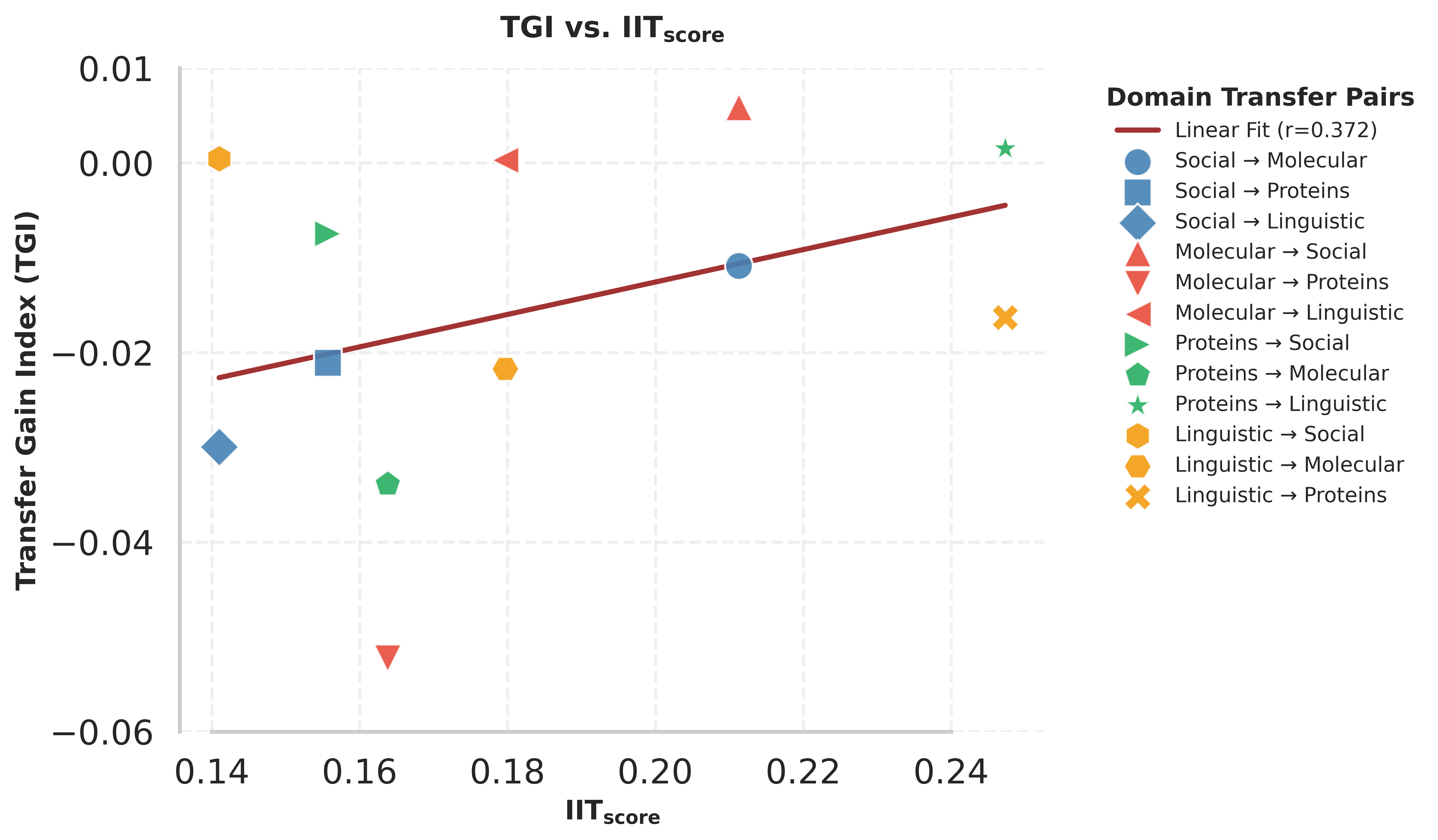}
    \caption{\textbf{Transfer Gain vs. Directed Affinity in High-Purity Regime.} Linear regression across 12 directed domain pairs ($\eta=0.1, \alpha=0.1$). The positive correlation ($r=0.372$) confirms that structural compatibility, quantified via the $\text{IIT}_{\text{score}}$, mitigate transfer degradation when absolute performance is constrained by near-perfect local baselines.}
    \label{fig:supp_reg_low_noise}
\end{figure}

\begin{figure}[H]
    \centering
    \includegraphics[width=0.65\textwidth]{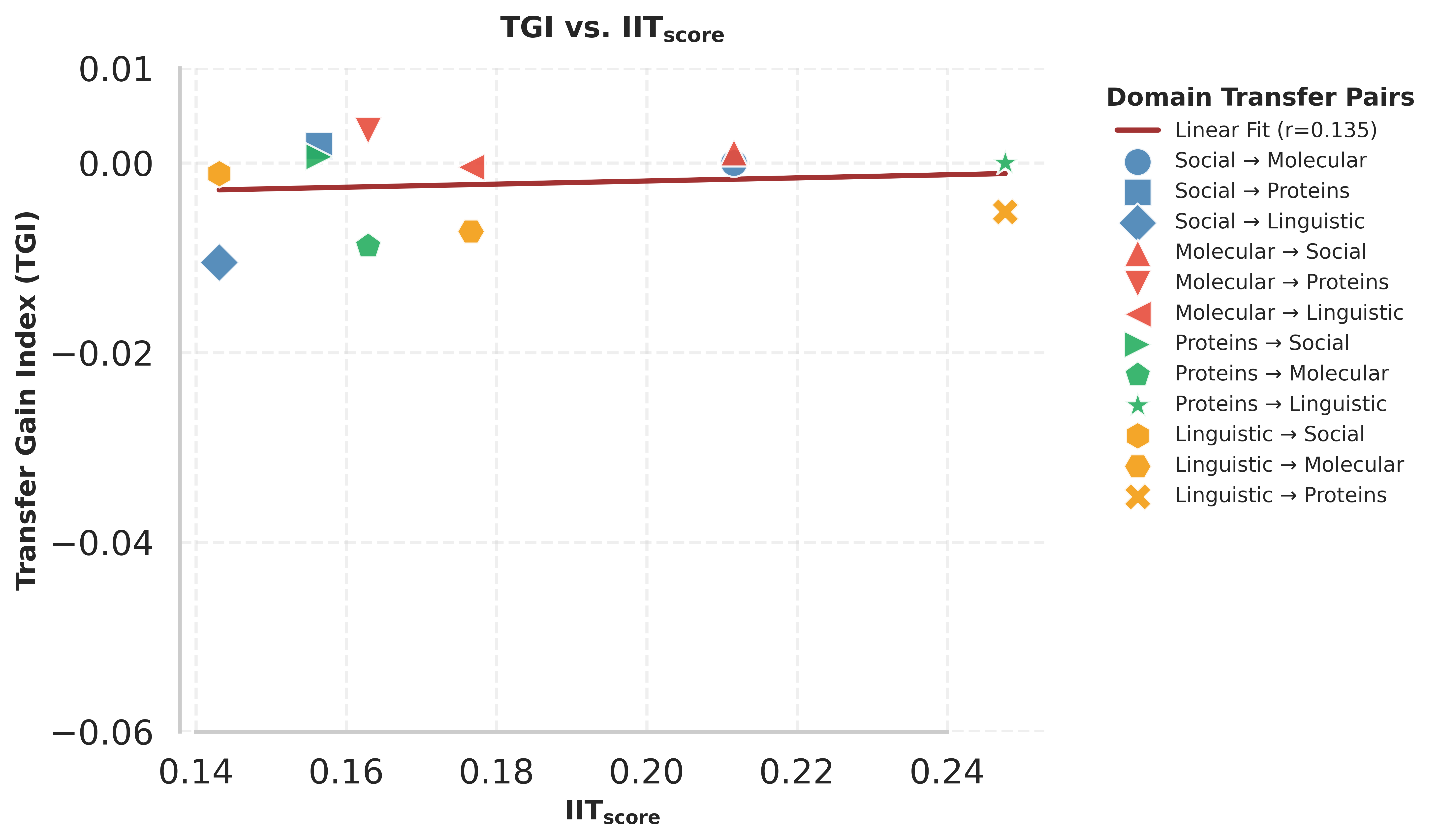} 
    \caption{\textbf{Transfer Gain vs. Directed Affinity in High-Noise Regime.} Linear regression across 12 directed domain pairs ($\eta=0.9, \alpha=0.9$). The correlation ($r=0.135$) indicates that even under maximum feature corruption, the stabilizing benefits of knowledge transfer are maximized for domain pairs exhibiting higher theoretical structural bridge strength.}
    \label{fig:supp_reg_high_noise}
\end{figure}

The association between the directed $\text{IIT}_{\text{score}}$ and the TGI identifies how theoretical manifold compatibility governs realized functional benefits. Regression analysis across the operational boundary conditions reveals consistent dynamics contingent upon signal purity:

\begin{enumerate}
    \item [] \textbf{High-Purity Mitigation ($r = 0.372$):} In regimes characterized by minimal noise ($\eta = 0.1$), structural affinity directly mitigates the performance decay typically induced by source-to-target sample imbalance (Fig.~\ref{fig:supp_reg_low_noise}). The significant positive correlation confirms that higher directed compatibility buffers the target manifold against distributional interference from the source domain, preserving the integrity of the local structural signal.
    \item [] \textbf{High-Noise Regularization ($r = 0.135$):} Under maximum feature corruption ($\eta = 0.9$), the predictive power of the $\text{IIT}_{\text{score}}$ remains positive, albeit attenuated (Fig.~\ref{fig:supp_reg_high_noise}). This trend identifies a \textit{rescue effect}, where knowledge propagation effectiveness is governed by the structural bridge strength. In these scenarios, an externally consistent source manifold acts as a rigid topological prosthesis, providing a stable discriminative grammar that pervasive noise in the target domain cannot obscure. The persistence of a positive correlation even under extreme stress validates the $\text{IIT}_{\text{score}}$ as a robust indicator of manifold compatibility.
\end{enumerate}

\bmhead{Acknowledgements}
This work was supported by the National Research Council (CNR), Italy. The author acknowledges the Institute of Cognitive Sciences and Technologies (ISTC-CNR) for providing the necessary research infrastructure. 

\bmhead{Funding}
This research was funded by the European Union - Next Generation EU - NRRP M6C2 - Investment 2.1 Enhancement and strengthening of biomedical research in the NHS and by FISM - Fondazione Italiana Sclerosi Multipla - cod. 2022/R-Multi/040 and financed or co-financed with the “5 per mille” public funding.

\bmhead{Competing Interests}
D.C. is a founding partner of AI2Life s.r.l. His primary employment is with the National Research Council (CNR), Italy. This work was conducted independently of AI2Life s.r.l. operational activities. The author declares no other competing financial or non-financial interests.

\bmhead{Ethics approval and consent to participate}
Not applicable. This study does not involve human participants, data, or tissue, nor does it involve any animal subjects.

\bmhead{Consent for publication}
Not applicable. This manuscript does not contain any individual person's data in any form (including individual details, images, or videos).

\bmhead{Data Availability}
The datasets analyzed during the current study (QM9, Facebook Ego, PROTEINS, Brown Corpus) are available in the respective public repositories cited in the Methods section. 

\bmhead{Materials availability}
Not applicable. No physical materials were generated or used in this study.

\bmhead{Code Availability}
The custom Python code used for the X-CDTL framework, the IIT mechanism, and all statistical analyses is available in a single comprehensive Jupyter Notebook on GitHub at \url{https://github.com/danielecaligiore/IIT_X-CDTL/tree/main}. 






\bibliography{sn-bibliography}

\end{document}